\documentclass{article}

\PassOptionsToPackage{numbers,sort&compress}{natbib}
\usepackage[main, final]{neurips_2026}
\usepackage{graphicx}
\usepackage{subcaption}
\usepackage{tikz}
\usepackage[rightcaption]{sidecap}

\usepackage[utf8]{inputenc} 
\usepackage[T1]{fontenc}    
\usepackage{hyperref}       
\usepackage{url}            
\usepackage{booktabs}       
\usepackage{amsfonts}       
\usepackage{nicefrac}       
\usepackage{microtype}      
\usepackage{xcolor}         
\usepackage{multirow}
\usepackage{amsmath,amssymb,amsthm,mathtools}
\usepackage{algorithm}
\usepackage{algpseudocode}
\usepackage{wrapfig}

\newtheorem{proposition}{Proposition}
\newtheorem{remark}{Remark}

\title{Robust Flow Matching under Target Corruption and Label Noise}

\author{%
  Mert Can Kurucu$^{1,2}$ \qquad
  Erik Englesson$^{3}$ \qquad
  Hossein Azizpour$^{1,4}$ \\[2mm]
  $^1$KTH Royal Institute of Technology, Stockholm, Sweden \\
  $^2$Istanbul Technical University, Istanbul, Türkiye \\
  $^3$University of Cambridge, Cambridge, United Kingdom \\[1mm]
  $^4$Science for Life Laboratory, Stockholm, Sweden \\[1mm]
  \texttt{kurucum@itu.edu.tr} \qquad
  \texttt{ebe23@cam.ac.uk} \qquad
  \texttt{azizpour@kth.se}
}

\begin{document}

\maketitle

\begin{abstract}
Flow Matching (FM) is a strong framework for generative modeling due to its stable training and efficient sampling. However, FM assumes clean target data and labels, an assumption often violated in practice. We formulate FM training as a time-dependent regression problem over conditional vector fields and analyze how the two corruption types affect the objective: i) target corruption adds a time-dependent residual term that grows with time, ii) label noise causes updates to the wrong class-conditional vector field. Based on this analysis, we propose to improve FM robustness by mapping corruption-specific residual scores to sample weights. For target corruption, we propose the \emph{Temporal Residual Score} (TRS), which emphasizes residual magnitudes at intermediate-to-late FM times, where clean and corrupted samples are more separable. For label noise, we propose the \emph{Comparative Conditional Score} (CCS), which compares the residual of the observed class-conditional vector field against alternative class-conditional vector fields for the same trajectory and velocity target. 
Experiments across controlled corruption settings and real-world noisy labels show that the proposed methods improve robustness over standard FM and robust-training baselines, supporting corruption-specific residual structure as a practical basis for robust FM. \footnote{Code is available at \url{https://github.com/mertcankurucu/robust-flow-matching}.}
\end{abstract}

\section{Introduction}
\label{sec:introduction}

Conditional generative models typically assume clean target data and correct conditioning labels. Yet, both assumptions can fail in practice, giving rise to two distinct corruption types: \emph{target corruption}, where the data samples to be modeled are corrupted, and \emph{conditional label noise}, where the conditioning label is incorrect. Target measurements may be corrupted or incomplete, and ``ground-truth'' labels may be imperfect surrogates affected by annotation error, reconstruction bias, or discretization. Consequently, generative performance can degrade substantially under target corruption and label noise, as observed in prior work on robust generative modeling \citep{daras2023ambient,thekumparampil2018rcgan}.

Flow Matching (FM) learns a time-dependent vector field (VF) that transports samples from a simple base distribution to the target data distribution \citep{lipman2023flow,chen2018neuralode,grathwohl2019ffjord,albergo2023building,liu2023rectifiedflow}. Although robustness has been extensively studied in supervised discriminative learning, it remains unclear how target corruption and label noise affect the time-dependent regression objective of FM and how training should be made robust. To our knowledge, this is the first FM-specific regression analysis of both corruption types, showing how their residual structure can be exploited for corruption-specific sample reweighting.

\begin{figure}
    \centering
    \includegraphics[width=\columnwidth]{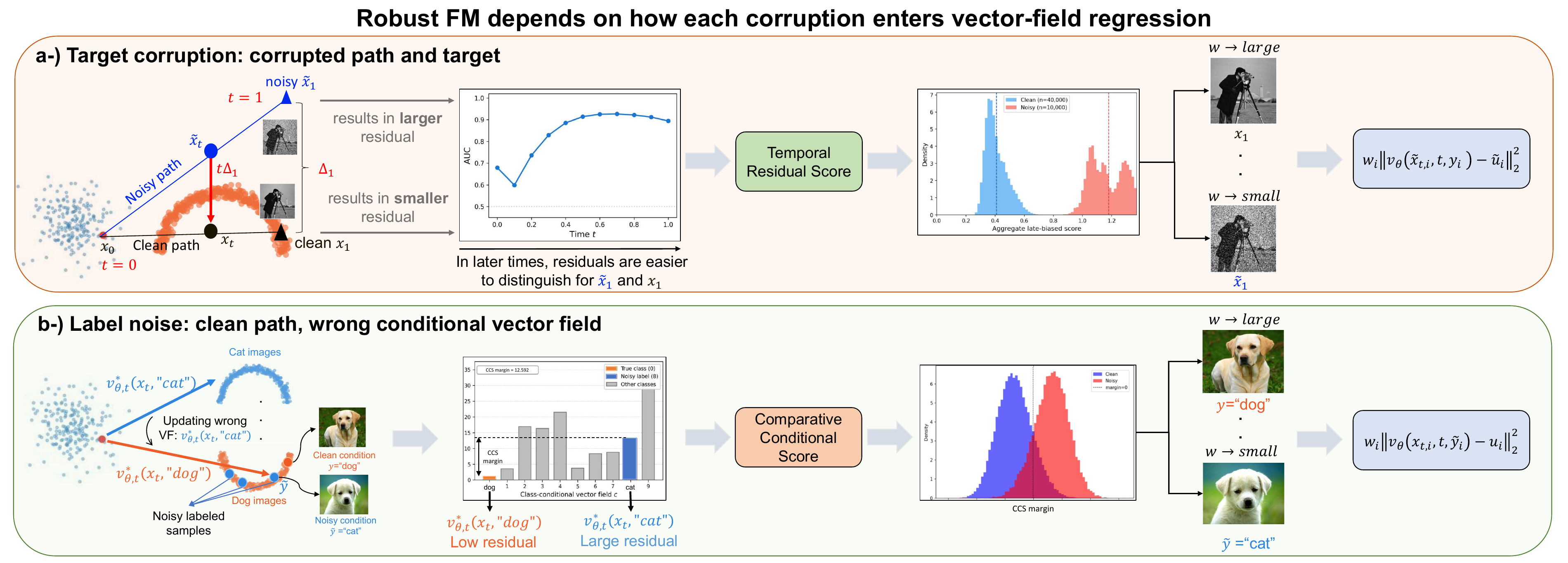}
    \caption{\textbf{Robust Flow Matching under target data corruption and label noise.}
    \textbf{(a)} Target corruption changes both the interpolation state and the velocity target,
    $\tilde{x}_t=x_t+t\Delta_1$ and $\tilde{u}=u+\Delta_1$, making residuals at
    intermediate-to-late FM times informative for assigning smaller weights to samples with stronger corruption effects.
    \textbf{(b)} Label noise leaves the path $(x_t,u)$ unchanged but routes the update
    through the wrong class-conditional vector field, motivating a comparison between
    the observed label and alternative class-conditional vector fields on the same
    trajectory.
    }
    \label{fig:main_figure}
\end{figure}

In this paper, we analyze and address FM under target corruption and label noise through the lens of robust regression over time-dependent conditional VFs. Our main contributions are:
\begin{itemize}
    \item \textbf{Theoretical analysis.} We formulate FM training as a regression problem over conditional VFs (Section~\ref{subsec:setup_cfm}) and show that the two corruptions enter differently: target corruption perturbs both the interpolation state and velocity target (Section~\ref{subsec:x_noise_in_fm}), whereas label noise leaves the interpolation state and velocity target unchanged but evaluates the same trajectory under an incorrect label, routing the update through the wrong class-conditional VF (Section~\ref{subsec:y_noise_in_fm}).

    \item \textbf{Robust FM training.} Our analysis identifies the residual structures that we use for corruption-specific sample reweighting under each corruption type. Target corruption inflates the residual through both a shifted velocity target and a state-perturbation term that grows with time, motivating the \emph{Temporal Residual Score} (TRS), which uses residual magnitude with late-biased time sampling (Section~\ref{subsec:method_x}). Label noise leaves the interpolation state and velocity target unchanged, so the residual increase arises from the mismatch of evaluating the clean target trajectory under the incorrect class-conditional VF. This motivates the \emph{Comparative Conditional Score} (CCS), which compares the residual of the observed class-conditional VF against alternative class-conditional VFs for the same trajectory (Section~\ref{subsec:method_y}). Both scores are mapped to soft sample weights in a weighted FM objective, without requiring clean data or an explicit noise model.

    \item \textbf{Experiments.}
    We evaluate target corruption and label noise on 2D benchmarks and CIFAR-10, and additionally evaluate label noise on CIFAR-100 and WebVision (Section~\ref{sec:experiments}). Further experiments examine larger condition spaces, and simultaneous target and label corruption. Comparisons with standard FM and adapted robust-training baselines demonstrate the benefit of corruption-specific residual scoring. Figure~\ref{fig:main_figure} provides an overview.
\end{itemize}

\section{Related Work}
\label{sec:related_work}

\paragraph{Robust supervised learning.} 
Robust learning under target data corruption and label noise has been extensively studied in supervised discriminative settings, including loss correction via transition matrix estimation \citep{patrini2017making,li2021provably,cheng2022classdependent}, sample selection, curriculum learning, and example reweighting \citep{han2018coteaching,jiang2018mentornet}, semi-supervised noise separation \citep{li2020dividemix}, robust losses \citep{englesson2021generalized,wang2019symmetric,ghosh2017robust}, confidence weighting \citep{berthon2021confidence}, and mixture-based robust regression \citep{lathuiliere2018deepgum}. These approaches limit the effect of unreliable supervision through correction, selection, reweighting, or robust penalties. Classical robust estimation, including Huber's M-estimators \citep{huber2009robust}, has also inspired robust losses and modern variants such as adaptive losses \citep{barron2019general} and curriculum-based truncation \citep{lyu2019curriculum}. Although they show that corrupted targets or labels can be handled through reweighting, correction, or selective training, they are designed for discriminative prediction or supervised regression rather than generative models.

\paragraph{Robust generative modeling.}
Robustness in generative modeling has been studied in several settings. One line learns clean target distributions from corrupted observations, starting from AmbientGAN~\citep{bora2018ambientgan} and extending to VAEs with incomplete data~\citep{mattei2019miwae}, normalizing flows for imputation~\citep{richardson2020mcflow}, and diffusion models trained with corrupted measurements~\citep{daras2023ambient,daras2023soft,daras2024tweedie,rozet2024learning,daras2025ambient}. Ambient Diffusion~\citep{daras2023ambient} assumes a known and applicable measurement-corruption operator, whereas our target corruption setting provides neither the corruption process nor sample-level corruption indicators. Another line studies conditional generation under noisy labels, including conditional GANs with label-transition structures~\citep{thekumparampil2018rcgan,kaneko2019labelnoise} and diffusion approaches such as TDSM~\citep{na2024labelnoise}, SBDC~\citep{cong2025sbdc}, and label-retrieval augmentation~\citep{chen2024labelretrieval}. Within FM, NRFlow~\citep{chen2025nrflow} studies a high-order formulation on synthetic distribution-transfer benchmarks rather than target-corrupted or label-noisy training data. Our work instead characterizes how corruptions enter FM regression and uses this residual structure for robust training.

\section{Robust Flow Matching under Target Corruption and Label Noise}
\label{sec:problem}

In this section, we formulate FM training as a regression problem in time-dependent conditional VFs, and show that target corruption and label noise enter this through different structured perturbations. 

\subsection{Conditional flow matching as time-dependent vector field regression}
\label{subsec:setup_cfm}

Let $\mathcal{X}\subseteq\mathbb{R}^d$ denote the target data space and $\mathcal{Y}$ the condition space, with clean data $(x_1,y)\sim p_{\mathrm{data}}(x_1,y)=p_{\mathrm{data}}(y)p_{\mathrm{data}}(x_1\mid y)$. 

We first analyze the two corruption channels separately: under target corruption, $x_1$ is corrupted while $y$ is clean, whereas label noise corrupts $y$ while $x_1$ is clean. In the target corruption setting, a clean pair $(x_1,y)$ is drawn from $p_{\mathrm{data}}$ and the endpoint is corrupted as $\tilde{x}_1\sim q_x(\cdot\mid x_1)$, yielding observed pairs $(\tilde{x}_1,y)\sim\tilde p_x$. Under label noise, the observed label is drawn as $\tilde y\sim q_y(\cdot\mid x_1,y)$, yielding pairs $(x_1,\tilde y)\sim\tilde p_y$; this includes class-dependent and instance-dependent noise. We assume no access to clean pairs, corruption indicators, or explicit models of $q_x$ or $q_y$. We also follow the classical robust-statistics view that corruption should be non-dominating: corrupted samples may be severe or arbitrary, but the clean component must remain statistically identifiable from the observed distribution~\citep{huber1964robust}. The operating conditions of the proposed scores are discussed in Appendix~\ref{app:limitations}. We additionally evaluate simultaneous corruption in Appendix~\ref{app:joint_corruption}.

Let $p_0$ be a simple source distribution on $\mathcal{X}$, chosen independently of $y$. For a clean pair $(x_1,y)\sim p_{\mathrm{data}}$ and source sample $x_0\sim p_0$, we consider conditional FM with the standard linear interpolation path $x_t=(1-t)x_0+tx_1$ for $t\in[0,1]$, and corresponding velocity target $u(x_0,x_1)=x_1-x_0$. CFM learns a time-dependent conditional VF $v_{\theta,t}(x, y) : \mathcal{X} \times \mathcal{Y} \to \mathbb{R}^d$ that defines the dynamics
\begin{equation}
\frac{d}{dt} x_t = v_{\theta,t}(x_t, y), \qquad x_0 \sim p_0,
\label{eq:cfm_ode}
\end{equation}
by regressing the model output to the velocity target. The clean CFM loss is
\begin{equation}
\mathcal{L}_{\mathrm{CFM}}(\theta)
=
\mathbb{E}_{\substack{(x_1, y) \sim p_{\mathrm{data}},\\ x_0 \sim p_0,\, t \sim \mathcal{U}(0,1)}}
\left[\bigl\|v_{\theta,t}(x_t, y) - u(x_0, x_1)\bigr\|_2^2\right].
\label{eq:cfm_clean_objective}
\end{equation}
This is a squared residual loss for the VF and therefore inherits the sensitivity of least-squares estimation to outliers~\citep{huber1964robust,huber2009robust}. The model takes $(x_t,t,y)$ as input and regresses onto the \emph{velocity target} $u(x_0,x_1)$, which is distinct from the \emph{target data sample} $x_1$ at the endpoint of the interpolation path. The key per-sample residual is
\begin{equation}
r_\theta^{\mathrm{clean}}(x_0, t; x_1, y) := v_{\theta,t}(x_t, y) - u(x_0, x_1),
\label{eq:cfm_clean_residual}
\end{equation}
so that $\mathcal{L}_{\mathrm{CFM}}(\theta) = \mathbb{E}\bigl[\|r_\theta^{\mathrm{clean}}\|_2^2\bigr]$. This residual view makes the connection to robust regression direct: corruptions can affect FM by changing the regression input $x_t$, the regression target $u(x_0,x_1)$, or the condition $y$ under which the vector field is evaluated. The next two sections show how target corruption and label noise affect this residual and the FM objective explicitly.

\subsection{Target corruption: corrupted path and shifted velocity target}
\label{subsec:x_noise_in_fm}

Under target corruption, the data sample $x_1$ is corrupted and the learner observes $\tilde{x}_1$. The noisy interpolation path and target velocity become $\tilde{x}_t=(1-t)x_0+t\tilde{x}_1$ and $\tilde{u}(x_0,\tilde{x}_1)=\tilde{x}_1-x_0$, giving the noisy FM loss
\begin{equation}
\tilde{\mathcal{L}}_{\mathrm{CFM}}^{x}(\theta)
=
\mathbb{E}_{\substack{(\tilde{x}_1,y)\sim \tilde{p}_x,\\ x_0\sim p_0,\, t\sim \mathcal{U}(0,1)}}
\left[\|v_{\theta,t}(\tilde{x}_t,y)-\tilde{u}(x_0,\tilde{x}_1)\|_2^2\right].
\label{eq:x_noise_objective}
\end{equation}
Thus, target corruption perturbs both the regression input $\tilde{x}_t$ and the regression target $\tilde{u}(x_0,\tilde{x}_1)$, but FM introduces additional structure because the perturbed input is an interpolation state whose corruption depends explicitly on time. Figure~\ref{fig:main_figure}(a) makes this geometry visible in 2D: the clean and corrupted trajectories start from the same source at $t=0$ and separate progressively along the path, with displacement $t\Delta_1$ that grows toward $t=1$. Proposition~\ref{prop:x_noise_decomposition} formalizes this picture.

\begin{proposition}
\label{prop:x_noise_decomposition}
Let $\Delta_1:=\tilde{x}_1-x_1$. Then
\begin{equation}
\tilde{x}_t=x_t+t\Delta_1,
\qquad
\tilde{u}(x_0,\tilde{x}_1)=u(x_0,x_1)+\Delta_1.
\label{eq:x_noise_exact_decomposition}
\end{equation}
Moreover, defining the noisy-target residual by
\begin{equation}
r_\theta^{x}(x_0,t;\tilde{x}_1,y)
:=
v_{\theta,t}(\tilde{x}_t,y)-\tilde{u}(x_0,\tilde{x}_1),
\label{eq:x_noise_residual_main}
\end{equation}
we obtain
\begin{equation}
r_\theta^{x}
=
r_\theta^{\mathrm{clean}}
+
\bigl(v_{\theta,t}(x_t+t\Delta_1,y)-v_{\theta,t}(x_t,y)\bigr)
-\Delta_1.
\label{eq:x_noise_residual_identity}
\end{equation}
\end{proposition}

Proposition~\ref{prop:x_noise_decomposition} shows that target corruption enters FM through two coupled terms (see Appendix~\ref{app:derivation_prop1} for derivation). The first is a \emph{state-perturbation term}, $v_{\theta,t}(x_t+t\Delta_1,y)-v_{\theta,t}(x_t,y)$, which measures how the conditional VF changes between the clean interpolation state $x_t$ and the corrupted state $x_t+t\Delta_1$. The second is a \emph{target-shift term} $-\Delta_1$, which reflects the direct perturbation of the regression target. Both vanish when $\Delta_1=0$ and grow with the magnitude of the endpoint corruption. Under the linear path, the discrepancy between the clean and corrupted states is $\tilde{x}_t-x_t=t\Delta_1$, so the state-perturbation term becomes more exposed as $t$ increases, making target corruption more visible at intermediate-to-late FM times. Further support is given in Appendix~\ref{app:diagnostics} and Figure~\ref{fig:trs_time_behavior}, where we visualize the time-dependent residual separation between clean and corrupted samples.

This also explains why residual magnitude at different time values becomes a useful corruption signal. In robust regression, corrupted samples often yield inflated residuals because the corruption directly affects the prediction error. Eq.~\eqref{eq:x_noise_residual_identity} shows that the same mechanism appears here, but with an FM-specific structure: $r_\theta^{x}$ is inflated not only by the shifted target $\tilde{u}(x_0,\tilde{x}_1)$, but also by the time-dependent discrepancy between $\tilde{x}_t$ and $x_t$. Once the model has fit clean samples well enough that $r_\theta^{\mathrm{clean}}$ is relatively small, these terms increase the residual for corrupted samples. This is the residual pattern used by TRS in Section~\ref{subsec:method_x}, and it suggests that in this setting, intermediate-to-late FM times should be more informative for separating clean and corrupted samples.

This residual pattern is not guaranteed for every corruption. TRS is intended for settings in which corrupted samples form a non-dominating component and produce intermediate-to-late-time residuals distinguishable from those of the clean majority. Rare valid modes or legitimate high-frequency structure may also yield large residuals and be downweighted, while corrupted samples close to the source distribution or otherwise easy to transport may yield small residuals and retain high weights. Appendix~\ref{app:limitations} discusses these operating conditions and limitations.

\subsection{Label noise: clean path and wrong conditional field}
\label{subsec:y_noise_in_fm}

Under label noise, the data sample $x_1$ remains clean while its conditioning label $y$ is replaced by $\tilde{y}\sim q_y(\cdot\mid x_1,y)$. Thus the interpolation path and target velocity remain $x_t=(1-t)x_0+tx_1$ and $u(x_0,x_1)=x_1-x_0$, and corruption enters only through the condition argument of the conditional VF. The noisy FM loss is
\begin{equation}
\tilde{\mathcal{L}}_{\mathrm{CFM}}^{y}(\theta)
=
\mathbb{E}_{\substack{(x_1,\tilde{y})\sim \tilde{p}_y,\\ x_0\sim p_0,\, t\sim \mathcal{U}(0,1)}}
\left[\|v_{\theta,t}(x_t,\tilde{y})-u(x_0,x_1)\|_2^2\right].
\label{eq:y_noise_objective}
\end{equation}
Figure~\ref{fig:main_figure}(b) shows the contrasting geometry: $(x_t,u)$ remains exactly the clean trajectory, but the update is routed through a different class-conditional VF, typically one corresponding to another valid class manifold rather than an arbitrary off-manifold direction. Label noise therefore carries structured directional information, not unstructured perturbation. Proposition~\ref{prop:y_noise_decomposition} makes this precise.

\begin{proposition}
\label{prop:y_noise_decomposition}
Define the noisy-condition residual by
\begin{equation}
r_\theta^{y}
:=
v_{\theta,t}(x_t,\tilde{y})-u(x_0,x_1).
\label{eq:y_noise_residual}
\end{equation}
Then
\begin{equation}
r_\theta^{y}
=
r_\theta^{\mathrm{clean}}
+
\Delta_{\theta,t}(x_t; y \to \tilde{y}),
\qquad
\Delta_{\theta,t}(x_t; y \to \tilde{y})
:=
v_{\theta,t}(x_t,\tilde{y})-v_{\theta,t}(x_t,y).
\label{eq:y_noise_exact_decomposition}
\end{equation}
Here, the notation $y \to \tilde y$ indicates that the same state $x_t$ at the same time $t$ is evaluated under the corrupted observed label $\tilde y$ instead of the clean label $y$.
\end{proposition}

Proposition~\ref{prop:y_noise_decomposition} shows that, under label noise, the corruption effect is concentrated entirely in the discrepancy between the observed and clean VFs (Appendix~\ref{app:derivation_prop2}). This decomposition has two consequences. First, residual magnitude can be informative because the observed conditional VF may be inconsistent with the clean transport target, which can increase the magnitude of $r_\theta^{y}$ in Eq.~\eqref{eq:y_noise_exact_decomposition}. However, a large residual can also arise from a difficult correctly labeled trajectory, so residual magnitude alone is not label-specific. Second, label noise has an additional comparative structure that is absent under target corruption. As illustrated in Figure~\ref{fig:main_figure}(b), the corruption does not push the trajectory to an arbitrary off-manifold location; but it evaluates the same clean trajectory under another valid condition, corresponding to a different class manifold. For a fixed $(x_t,t)$ and target velocity $u(x_0,x_1)$, the alternative class-conditional VFs $\{v_{\theta,t}(\cdot,c): c\in\mathcal{Y},\, c\neq \tilde y\}$ provide alternative explanations of the same trajectory. Thus, the relevant question is not only whether the residual under $\tilde y$ is large, but whether some alternative label $c\neq \tilde y$ explains the trajectory better. This comparative structure motivates CCS in Section~\ref{subsec:method_y}.

\section{Robust Flow Matching with Corruption-Specific Sample Reweighting}
\label{sec:method}

The analysis in Section~\ref{sec:problem} identifies, for each corruption type, the residual structure that we use to assign sample weights. Both methods retain the per-sample squared-residual FM loss and modulate each sample by a weight $w_i\in[0,1]$ computed from a corruption-specific score. We optimize
\begin{equation}
\mathcal{L}_{\mathrm{WFM}}(\theta)
=
\frac{1}{N}\sum_{i=1}^{N}\mathbb{E}_{t\sim\pi}\bigl[w_i\,\ell_i(\theta;t)\bigr],
\label{eq:method_weighted_objective}
\end{equation}
where $\ell_i(\theta;t)$ is the per-sample FM loss of sample $i$ at time $t$ and $\pi$ is the time-sampling distribution used by the corresponding method. We convert corruption scores to sample weights using a smooth monotone map,
$w_i=\omega(s_i;q,\tau):=\sigma\!\left(-\frac{s_i-q}{\tau}\right)$, where $\sigma$ is the logistic sigmoid, $q$ is the transition center, and $\tau>0$ controls the sharpness of the transition. Thus, low-score samples receive larger weights, high-score samples receive smaller weights. The question is how to construct the score $s_i$.

Both methods use sample-indexed scores to determine soft sample weights, but differ in how the scores are updated. FM-TRS stores one residual score per sample in a buffer $S\in\mathbb{R}^N$. Whenever a sample appears in a minibatch, its instantaneous residual score is incorporated into the stored score using an exponential moving average. FM-CCS instead recomputes comparative scores at scheduled refresh points, maps them to soft weights, and keeps these weights fixed until the next refresh. Its comparisons across all labels are therefore performed only at refresh points, rather than at every training step. In both methods, each sample weight multiplies the corresponding per-sample FM loss in Eq.~\eqref{eq:method_weighted_objective}. The score definitions and weighting rules are detailed in Sections~\ref{subsec:method_x} and~\ref{subsec:method_y}, while Algorithms~\ref{alg:x_noise} and~\ref{alg:y_noise} summarize the complete training procedures.

\subsection{Temporal residual score for target corruption}
\label{subsec:method_x}

Section~\ref{subsec:x_noise_in_fm} identifies how target corruption changes the FM residual through the displaced interpolation state and shifted velocity target. The observed separation between clean and corrupted residuals at intermediate-to-late FM times motivates a residual-based score with late-biased time sampling.

For a sampled tuple $(x_0,t,\tilde{x}_1,y)$, we define the instantaneous target corruption score
\begin{equation}
s_\theta^{x}(x_0,t;\tilde{x}_1,y)
:=
\|r_\theta^{x}(x_0,t;\tilde{x}_1,y)\|_2^2,
\label{eq:method_x_instantaneous_score}
\end{equation}
where $r_\theta^{x}$ is the noisy-target residual in Eq.~\eqref{eq:x_noise_residual_main}. This is the per-sample FM loss under target corruption and requires no additional model evaluation. We use the late-biased time distribution
\begin{equation}
\pi_x(t)=2t,\qquad t\in[0,1],
\label{eq:method_x_time_dist}
\end{equation}
implemented by $t=\sqrt{\xi}$ with $\xi\sim\mathcal{U}(0,1)$. This choice biases score computation toward FM times where the corrupted-state discrepancy $x_t+t\Delta_1$ is more exposed.

The same sampled time and residual are used both to update TRS and to compute the weighted FM training loss. Reusing this residual avoids an additional forward pass for an independently sampled scoring time. Consequently, late-biased sampling affects both score accumulation and optimization. Since
$\mathbb{E}_{t\sim\pi_x}[\ell_i(\theta;t)]
=\mathbb{E}_{t\sim\mathcal{U}(0,1)}[2t\,\ell_i(\theta;t)]$,
it introduces a temporal weight $2t$ in addition to sample reweighting. Table~\ref{tab:abl_trs_timebias_only} separates the effect of this temporal weighting from TRS, and Appendix~\ref{app:clean_time_sampling} evaluates its effect on clean-data generation.

A single residual evaluation can be a noisy estimate of sample reliability, since it depends on the sampled source point $x_0$, the sampled FM time $t$, minibatch stochasticity, and the current state of the model during training. We therefore aggregate residual evidence online using a sample-indexed exponential moving average (EMA). Let $S_i^{(k)}$ be the stored score of sample $i$ after step $k$, and let $\beta\in[0,1)$ be the EMA decay. If sample $i$ appears in the minibatch at step $k$ with $t_k\sim\pi_x$, we update
\begin{equation}
S_i^{(k)}
=
\beta S_i^{(k-1)}
+
(1-\beta)\,
s_\theta^{x}(x_{0,i}^{(k)},t_k;\tilde{x}_{1,i}^{(k)},y_i^{(k)}).
\label{eq:method_x_ema}
\end{equation}
Otherwise, we keep $S_i^{(k)}=S_i^{(k-1)}$. We refer to $S_i^{(k)}$ as the \emph{Temporal Residual Score} (TRS). All samples are scored under the same time distribution $\pi_x$, and the EMA reduces the influence of any single stochastic residual evaluation. Since the scale of $S_i^{(k)}$ changes during training, we center the weighting rule at the empirical $(1-\rho_x)$-quantile of the current score buffer,
$q_x^{(k)}=Q_{1-\rho_x}(\{S_j^{(k)}\}_{j=1}^{N})$,
and set
$w_i^{x,(k)}=\omega(S_i^{(k)};q_x^{(k)},\tau_x)$.
Thus, $\rho_x$ controls the high-score tail affected by reweighting without requiring an absolute residual threshold. It is a robustness hyperparameter, not an assumed corruption rate. During an initial warmup, we accumulate scores while keeping all sample weights equal to one, allowing the VF to begin fitting the data before reweighting is applied. Appendix~\ref{app:ablations} evaluates warmup, EMA decay, quantile centering, and time sampling.

\subsection{Comparative conditional score for label noise}
\label{subsec:method_y}

Under label noise, Section~\ref{subsec:y_noise_in_fm} showed that the interpolation state $x_t$ and target velocity $u(x_0,x_1)$ remain clean, and that corruption acts only through the conditional VF $v_{\theta,t}(x_t,\tilde{y})$. Residual magnitude $\|r_\theta^y\|$ can still be informative, but a large residual may arise from an incorrect label or a difficult correctly labeled trajectory. Label noise also provides a comparative signal: on the same clean trajectory, the correct class-conditional VF should yield a lower residual than alternative class-conditional VFs. For a sampled tuple $(x_0,t,x_1,\tilde y)$, define the label-specific squared residual
\begin{equation}
\ell_\theta^{y}(x_0,t,x_1;c)
:=
\|v_{\theta,t}(x_t,c)-u(x_0,x_1)\|_2^2,
\label{eq:method_y_per_class_loss}
\end{equation}
for any $c\in\mathcal{Y}$; for $c=\tilde y$, this recovers $\|r_\theta^y\|_2^2$. This quantity is a per-label FM loss, not yet a corruption score. We then define the instantaneous comparative label noise score as
\begin{equation}
s_\theta^{y}(x_0,t,x_1,\tilde y)
:=
\ell_\theta^{y}(x_0,t,x_1;\tilde y)
-
\min_{c\neq \tilde y}\ell_\theta^{y}(x_0,t,x_1;c).
\label{eq:method_y_instantaneous_score}
\end{equation}
Positive values indicate that an alternative class-conditional VF explains the same trajectory better than the observed one, and therefore provide evidence that the observed label may be corrupted. Negative values indicate that the observed conditional VF explains the trajectory better than every alternative. We use this score only for soft weighting, not for hard relabeling.

To reduce variance, we average this score over a fixed deterministic set of probe times $\mathcal{T}_y\subset[0,1]$:
\begin{equation}
M_i =
\frac{1}{|\mathcal{T}_y|}
\sum_{t\in \mathcal{T}_y}
s_\theta^{y}(x_{0,i},t,x_{1,i},\tilde y_i).
\label{eq:method_y_ccs}
\end{equation}
We refer to $M_i$ as the Comparative Conditional Score (CCS). Each comparison uses the same state, time, and velocity target for all labels; Eq.~\eqref{eq:method_y_ccs} averages these within-time margins rather than directly comparing raw losses from different times. An ablation on the size and composition of different sets are given in Table~\ref{tab:abl_ccs_probes}.

During ordinary FM updates under label noise, we use the standard uniform time distribution $\pi_y=\mathcal{U}(0,1)$; the fixed probe set $\mathcal{T}_y$ is used only for refreshing CCS weights. CCS is not evaluated continuously from the start of training. Early on, class-conditional VFs may not yet be sufficiently separated, and comparing alternative labels is expensive. We therefore refresh CCS only a few times; in our experiments, we refresh it at $25\%$ and $50\%$ of training (ablations in Appendix~\ref{app:ablations}, Table~\ref{tab:abl_ccr_refresh}).

The computational-cost analysis in Appendix~\ref{app:computational_cost} and Table~\ref{tab:cost_main} shows that this schedule adds only a small computational overhead. An exact refresh requires $O(N|\mathcal{T}_y|C)$ sample--condition evaluations, where $C=|\mathcal{Y}|$. Although comparisons are vectorized, their practical cost grows with the number of conditions and model complexity.

Unlike TRS, CCS has a natural decision boundary at zero, so it does not require quantile centering. We map it to weights through $w_i^y
=
\omega(M_i;0,\tau_y)
=
\sigma\!\left(-\frac{M_i}{\tau_y}\right)$, where $\tau_y$ controls the sharpness of the transition around zero. Proposition~\ref{prop:ccs_sign} in Appendix~\ref{app:proofs} gives a sufficient class-conditional margin condition under which clean and corrupted observed labels fall on opposite sides of this zero-centered boundary. Specifically, the condition requires the correct-label loss to be lower than every competing-label loss by a positive margin at each probe time. This concerns the evaluated trajectories and probe times, rather than global separation of the conditional VFs. Figure~\ref{fig:score_behavior} provides an empirical diagnostic. Larger scores receive smaller weights, while negative scores approach unit weight as their magnitude grows relative to $\tau_y$. Pseudocode for both TRS and CCS is given in Appendix~\ref{app:pseudocode}.

\section{Experiments}
\label{sec:experiments}

We evaluate FM-TRS and FM-CCS on 2D benchmarks commonly used in flow-based generative modeling~\citep{lipman2023flow,albergo2023building}, CIFAR-10 and CIFAR-100~\citep{krizhevsky2009learning}, and WebVision~\citep{li2017webvision}. The 2D benchmarks expose the geometry of the two corruption mechanisms, CIFAR-10 tests image-scale robustness under controlled corruptions, CIFAR-100 extends the label-noise evaluation to more classes and a larger model, and WebVision provides a real-world noisy-label evaluation. The experiments test whether \textbf{(i)} the proposed reweighting methods mitigate degradation under both target corruption and label noise, and \textbf{(ii)} the FM-specific design improves over generic robust-regression and noisy-label baselines that do not use the structure of the FM objective. Results are mean $\pm$ standard deviation over three seeds; paired seed-wise significance tests for the main 2D, CIFAR-10, and WebVision comparisons are in Appendix~\ref{app:statistical_significance}. We use corruption indicators only for diagnostics, showing that TRS and CCS separate clean from corrupted samples without using these indicators during training; see Appendix~\ref{app:diagnostics}. Implementation details, corruption procedures, hyperparameters, baseline adaptations, and ablations are given in Appendices~\ref{app:implementation} and~\ref{app:ablations}. For the main controlled experiments on 2D benchmarks, CIFAR-10, and CIFAR-100, we report two temperatures,
$\tau\in\{0.001,0.01\}$, corresponding to different sharpness levels in the
score-to-weight map $\omega$ defined in Section~\ref{sec:method}; for WebVision,
we use $\tau_y=0.001$.

\paragraph{Baselines.}
We compare with standard optimal transport FM (OT-FM) \citep{lipman2023flow}, robust-loss variants using L1 or Huber ($\delta=1.0$), and FM adaptations of DeepGUM \citep{lathuiliere2018deepgum} and DivideMix (co-divide only) \citep{li2020dividemix}, where applicable. Full implementation details and the design choices made when porting DeepGUM and DivideMix to FM are given in Appendix~\ref{app:adapted_baselines}. All the baselines test whether robustness can be obtained by directly importing generic robust-regression or noisy-label methods into FM, without exploiting the structure analyzed in Secs.~\ref{subsec:x_noise_in_fm}--\ref{subsec:y_noise_in_fm}. Additional qualitative samples are in Appendix~\ref{app:additional_samples}. 

\subsection{2D benchmarks}
\label{subsec:exp_2d}

We use 2D benchmarks to make the geometry of target corruption and label noise directly visible, following common practice in FM and related flow-based generative modeling work~\citep{lipman2023flow,albergo2025stochastic}. For target corruption, we report Wasserstein-2 distance ($W_2$), Maximum Mean Discrepancy (MMD), and Coverage against the clean target distribution; for label noise, we report mean class-wise (CW) $W_2$ and CW-Coverage~\citep{na2024labelnoise}. Metric definitions are given in Appendix~\ref{app:metrics}.

\begin{figure}[!t]
    \centering
    \includegraphics[width=\linewidth]{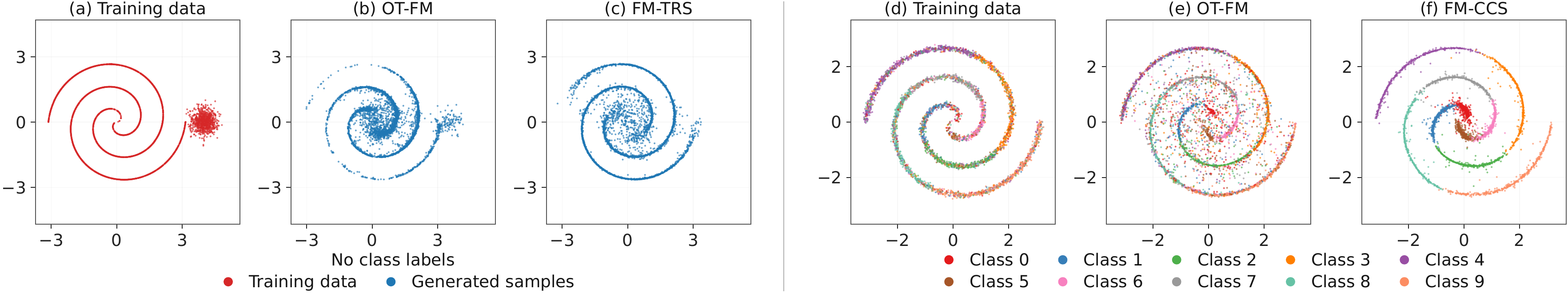}
\caption{\textbf{2D results.} (a)-(c) Under target corruption, FM-TRS downweights the off-manifold outliers; (d)-(f) under label noise, FM-CCS reduces cross-class mixing and improves class-aligned spirals.}
    \label{fig:app_2d_spirals_ynoise}
\end{figure}

\begin{table}[!t]
\centering
\caption{\textbf{2D target corruption results for spirals.}
FM-TRS improves over robust-regression baselines under off-manifold target corruption. ``Uniform'' denotes TRS scoring with uniform rather than late-biased time sampling, and bold numbers indicate the best performance in each column.}
\label{tab:2d_xnoise_shifted_spirals}
\scriptsize
\setlength{\tabcolsep}{3.5pt}
\renewcommand{\arraystretch}{0.90}
\begin{tabular}{lccc}
\toprule
Method & $W_2 \downarrow$ & MMD $\downarrow$ & Coverage $\uparrow$ \\
\midrule
OT-FM & $1.199{\scriptstyle\pm0.097}$ & $0.279{\scriptstyle\pm0.033}$ & $0.861{\scriptstyle\pm0.018}$ \\
OT-FM (L1) & $1.164{\scriptstyle\pm0.022}$ & $0.247{\scriptstyle\pm0.006}$ & $0.872{\scriptstyle\pm0.002}$ \\
OT-FM (Huber) & $1.239{\scriptstyle\pm0.091}$ & $0.281{\scriptstyle\pm0.032}$ & $0.864{\scriptstyle\pm0.021}$ \\
FM-DeepGUM & $0.727{\scriptstyle\pm0.018}$ & $0.129{\scriptstyle\pm0.007}$ & $0.888{\scriptstyle\pm0.003}$ \\
FM-TRS Uniform $\tau{=}0.001$ & $0.621{\scriptstyle\pm0.030}$ & $0.093{\scriptstyle\pm0.010}$ & $0.891{\scriptstyle\pm0.003}$ \\
FM-TRS Uniform $\tau{=}0.01$ & $0.617{\scriptstyle\pm0.038}$ & $0.093{\scriptstyle\pm0.012}$ & $0.895{\scriptstyle\pm0.004}$ \\
\midrule
FM-TRS $\tau{=}0.001$ & $0.532{\scriptstyle\pm0.019}$ & $0.067{\scriptstyle\pm0.006}$ & $0.895{\scriptstyle\pm0.002}$ \\
FM-TRS $\tau{=}0.01$ & {\bfseries\boldmath $0.434{\scriptstyle\pm0.011}$} & {\bfseries\boldmath $0.048{\scriptstyle\pm0.007}$} & {\bfseries\boldmath $0.897{\scriptstyle\pm0.002}$} \\
\bottomrule
\end{tabular}
\end{table}

\begin{table}[!t]
\centering
\caption{\textbf{2D label-noise results on segmented spirals.} CW-$W_2$ and CW-Cov.\ are class-wise metrics averaged over classes, and bold numbers indicate the best performance in each column.}
\label{tab:2d_ynoise_classw2}
\scriptsize
\setlength{\tabcolsep}{3.5pt}
\renewcommand{\arraystretch}{0.90}
\begin{tabular}{lcccccc}
\toprule
& \multicolumn{2}{c}{Sym.\ 0.2} & \multicolumn{2}{c}{Sym.\ 0.4} & \multicolumn{2}{c}{Asym.\ 0.4} \\
\cmidrule(lr){2-3}\cmidrule(lr){4-5}\cmidrule(lr){6-7}
Method & CW-$W_2 \downarrow$ & CW-Cov.\ $\uparrow$ & CW-$W_2 \downarrow$ & CW-Cov.\ $\uparrow$ & CW-$W_2 \downarrow$ & CW-Cov.\ $\uparrow$ \\
\midrule
OT-FM
& $1.018{\scriptstyle\pm0.029}$ & $0.678{\scriptstyle\pm0.028}$
& $1.419{\scriptstyle\pm0.036}$ & $0.613{\scriptstyle\pm0.021}$
& $1.443{\scriptstyle\pm0.037}$ & $0.666{\scriptstyle\pm0.014}$ \\
OT-FM (Huber)
& $0.826{\scriptstyle\pm0.045}$ & $0.700{\scriptstyle\pm0.016}$
& $1.276{\scriptstyle\pm0.051}$ & $0.662{\scriptstyle\pm0.014}$
& $1.331{\scriptstyle\pm0.036}$ &  $0.673{\scriptstyle\pm0.032}$ \\
OT-FM (L1)
& $0.743{\scriptstyle\pm0.039}$ & $0.708{\scriptstyle\pm0.026}$
& $1.195{\scriptstyle\pm0.040}$ & $0.654{\scriptstyle\pm0.026}$
& $1.258{\scriptstyle\pm0.037}$ & $0.671{\scriptstyle\pm0.014}$ \\
FM-DivideMix
& $0.368{\scriptstyle\pm0.022}$ & $0.693{\scriptstyle\pm0.006}$
& $0.522{\scriptstyle\pm0.024}$ & $0.688{\scriptstyle\pm0.019}$
& $1.009{\scriptstyle\pm0.024}$ & $0.563{\scriptstyle\pm0.020}$ \\
\midrule
FM-CCS $\tau{=}0.001$
& $0.183{\scriptstyle\pm0.003}$ & $0.700{\scriptstyle\pm0.014}$
& $0.194{\scriptstyle\pm0.013}$ & {\bfseries\boldmath $0.725{\scriptstyle\pm0.033}$}
& {\bfseries\boldmath $0.832{\scriptstyle\pm0.032}$} & $0.674{\scriptstyle\pm0.029}$ \\
FM-CCS $\tau{=}0.01$
& {\bfseries\boldmath $0.182{\scriptstyle\pm0.003}$} & {\bfseries\boldmath $0.710{\scriptstyle\pm0.019}$}
& {\bfseries\boldmath $0.188{\scriptstyle\pm0.006}$} & $0.722{\scriptstyle\pm0.032}$
& $0.834{\scriptstyle\pm0.008}$ & {\bfseries\boldmath $0.677{\scriptstyle\pm0.024}$} \\
\bottomrule
\end{tabular}%

\end{table}

Table~\ref{tab:2d_xnoise_shifted_spirals} shows that, compared with OT-FM, FM-TRS reduces $W_2$ from $1.199$ to $0.434$ and MMD from $0.279$ to $0.048$. Generic robust regression helps only partially: FM-DeepGUM improves over L1 and Huber losses, but FM-TRS still reduces $W_2$ and MMD by about $40\%$ and $63\%$ relative to FM-DeepGUM. The late-biased TRS variant also improves over uniform-time TRS, supporting the role of FM-time structure in residual-based reweighting. Figure~\ref{fig:app_2d_spirals_ynoise}(a)--(c) shows the same trend qualitatively: OT-FM is pulled toward the off-manifold noise, whereas FM-TRS suppresses it.

Table~\ref{tab:2d_ynoise_classw2} shows that FM-CCS gives the best CW-$W_2$ across all label-noise regimes. Relative to OT-FM, the best CCS variant reduces CW-$W_2$ by about $82\%$, $87\%$, and $42\%$ under symmetric $20\%$, symmetric $40\%$, and asymmetric $40\%$ noise, respectively. It also improves over FM-DivideMix by about $51\%$, $64\%$, and $18\%$, showing that generic noisy-label selection is less effective than comparing alternative class-conditional VFs. Figure~\ref{fig:app_2d_spirals_ynoise}(d)--(f) illustrates this effect: OT-FM exhibits cross-class mixing, whereas FM-CCS produces cleaner class-aligned spirals.

\paragraph{Additional corruption settings.}
CCS also improves CW-$W_2$ and CW-Coverage under $60\%$ symmetric and $40\%$ instance-dependent label noise (Appendix~\ref{app:complex_label_noise}). We also evaluate simultaneous target corruption and label noise using the combined weights $w_i^{\mathrm{joint}}=w_i^x w_i^y$. On segmented spirals with $20\%$ target outliers and $20\%$ or $40\%$ label flips, this reduces CW-$W_2$ by approximately $88\%$ and $86\%$, respectively, relative to OT-FM (Appendix~\ref{app:joint_corruption}).

\subsection{CIFAR-10 under target corruption}
\label{subsec:exp_cifar_xnoise}

\begin{figure}[!t]
    \centering

    \begin{minipage}{0.46\textwidth}
        \centering
        \textbf{Target corruption}
    \end{minipage}
    \hspace{0.04\textwidth}
    \begin{minipage}{0.46\textwidth}
        \centering
        \textbf{Label noise}
    \end{minipage}

    \vspace{2pt}

    \begin{subfigure}[t]{0.225\textwidth}
        \centering
        \includegraphics[width=\linewidth]{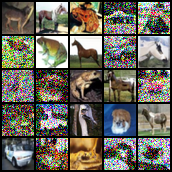}
        \caption*{(a) OT-FM}
    \end{subfigure}\hfill
    \begin{subfigure}[t]{0.225\textwidth}
        \centering
        \includegraphics[width=\linewidth]{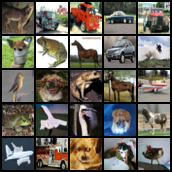}
        \caption*{(b) FM-TRS}
    \end{subfigure}
    \hspace{0.04\textwidth}
    \begin{subfigure}[t]{0.225\textwidth}
        \centering
        \includegraphics[width=\linewidth]{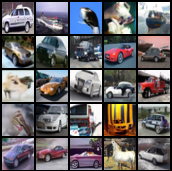}
        \caption*{(c) OT-FM}
    \end{subfigure}\hfill
    \begin{subfigure}[t]{0.225\textwidth}
        \centering
        \includegraphics[width=\linewidth]{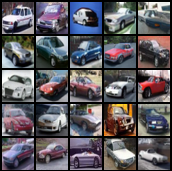}
        \caption*{(d) FM-CCS}
    \end{subfigure}

    \caption{\textbf{Qualitative CIFAR-10 results under target corruption and label noise.}
    \textbf{(a)--(b)} Under target corruption, FM-TRS suppresses visible corruption
    artifacts and produces cleaner samples than OT-FM.
    \textbf{(c)--(d)} Under label noise, the main failure mode is class misalignment:
    for the car class, OT-FM often drifts toward other categories, whereas FM-CCS
    preserves stronger class consistency.}
    \label{fig:cifar_main}
\end{figure}

For CIFAR-10 target corruption, we use the standard FM UNet setup from~\citep{lipman2023flow} and corrupt $20\%$ of training images with either Gaussian or salt-and-pepper (SP) noise. This setting follows the robust-regression regime considered in Section~\ref{subsec:setup_cfm}: corrupted targets are severe but non-dominating, so the clean distribution remains identifiable, and the two noise types test both dense and sparse image corruptions. Table~\ref{tab:cifar10_xnoise} shows that OT-FM is severely degraded under target corruption: FID increases from $2.67$ to $73.17$ under Gaussian corruption and to $64.73$ under SP corruption. FM-TRS reduces these values to $2.91$ and $3.61$, respectively, nearly recovering the clean reference. Generic robust-regression baselines help only partially: OT-FM (L1) improves over OT-FM, OT-FM (Huber) remains close to OT-FM, and FM-DeepGUM is stronger than both, but our FM-TRS still achieves better FID under Gaussian corruption ($2.91$ vs. $4.43$) and SP corruption ($3.61$ vs. $4.11$). Figure~\ref{fig:cifar_main}(a)--(b) shows the same effect qualitatively: OT-FM reproduces visible corruption artifacts, whereas FM-TRS suppresses them. Overall, exploiting the FM-specific corruption structure yields stronger robustness than generic robust-regression baselines.

\begin{table}[!t]
\centering
\caption{\textbf{CIFAR-10 target corruption results.}
We corrupt training images with Gaussian or Salt-and-Pepper noise and evaluate against the clean test distribution. ``Uniform'' denotes TRS scoring with uniform rather than late-biased time sampling, and bold numbers indicate the best performance.}
\label{tab:cifar10_xnoise}
\scriptsize
\setlength{\tabcolsep}{3.5pt}
\renewcommand{\arraystretch}{0.90}
\begin{tabular}{lcccccc}
\toprule
& \multicolumn{3}{c}{Gaussian} & \multicolumn{3}{c}{Salt \& Pepper} \\
\cmidrule(lr){2-4}\cmidrule(lr){5-7}
Method & FID $\downarrow$ & IS $\uparrow$ & Cov $\uparrow$
       & FID $\downarrow$ & IS $\uparrow$ & Cov $\uparrow$ \\
\midrule
Clean FM
& $2.67\scriptstyle\pm 0.03$ & $9.53\scriptstyle\pm 0.06$ & $0.971\scriptstyle\pm 0.002$
& $2.67\scriptstyle\pm 0.03$ & $9.53\scriptstyle\pm 0.06$ & $0.971\scriptstyle\pm 0.002$ \\
\midrule
OT-FM
& $73.17\scriptstyle\pm 0.69$ & $6.26\scriptstyle\pm 0.05$ & $0.899\scriptstyle\pm 0.001$
& $64.73\scriptstyle\pm 1.41$ & $7.41\scriptstyle\pm 0.06$ & $0.910\scriptstyle\pm 0.003$ \\
OT-FM (L1)
& $24.56\scriptstyle\pm 1.39$ & $8.65\scriptstyle\pm 0.06$ & $0.921\scriptstyle\pm 0.004$
& $15.82\scriptstyle\pm 0.69$ & $7.69\scriptstyle\pm 0.05$ & $0.927\scriptstyle\pm 0.003$ \\
OT-FM (Huber)
& $67.02\scriptstyle\pm 0.77$ & $6.56\scriptstyle\pm 0.01$ & $0.910\scriptstyle\pm 0.003$
& $58.46\scriptstyle\pm 1.02$ & $7.71\scriptstyle\pm 0.07$ & $0.919\scriptstyle\pm 0.003$ \\
FM-DeepGUM
& $4.43\scriptstyle\pm 0.19$ & $8.88\scriptstyle\pm 0.03$ & $0.949\scriptstyle\pm 0.002$
& $4.11\scriptstyle\pm 0.19$ & $8.91\scriptstyle\pm 0.05$ & $0.950\scriptstyle\pm 0.004$ \\
FM-TRS (uniform), $\tau{=}0.001$
& $3.25\scriptstyle\pm 0.11$ & $9.19\scriptstyle\pm 0.02$ & $0.968\scriptstyle\pm 0.001$
& $3.75\scriptstyle\pm 0.18$ & $9.25\scriptstyle\pm 0.02$ & $0.964\scriptstyle\pm 0.001$ \\
FM-TRS (uniform), $\tau{=}0.01$
& $3.36\scriptstyle\pm 0.12$ & $9.17\scriptstyle\pm 0.07$ & $0.964\scriptstyle\pm 0.001$
& $3.71\scriptstyle\pm 0.24$ & $9.24\scriptstyle\pm 0.03$ & $0.964\scriptstyle\pm 0.003$ \\
\midrule
FM-TRS, $\tau{=}0.001$
& $2.94\scriptstyle\pm 0.13$ & {\bfseries\boldmath $9.35\scriptstyle\pm 0.01$} & $0.965\scriptstyle\pm 0.002$
& {\bfseries\boldmath $3.61\scriptstyle\pm 0.26$} & {\bfseries\boldmath $9.47\scriptstyle\pm 0.08$} & $0.965\scriptstyle\pm 0.003$ \\
FM-TRS, $\tau{=}0.01$
& {\bfseries\boldmath $2.91\scriptstyle\pm 0.09$} & $9.33\scriptstyle\pm 0.10$ & {\bfseries\boldmath $0.969\scriptstyle\pm 0.002$}
& $4.36\scriptstyle\pm 0.04$ & $9.46\scriptstyle\pm 0.05$ & {\bfseries\boldmath $0.968\scriptstyle\pm 0.002$} \\
\bottomrule
\end{tabular}%

\end{table}

\begin{table}[!t]
\centering
\caption{\textbf{CIFAR-10 label-noise results.}
Sym. and Asym. denote symmetric and asymmetric label noise, with headers indicating noise rate. CW-FID and CW-Cov. are metrics averaged over classes.}
\label{tab:cifar10_ynoise_extended}
\scriptsize
\setlength{\tabcolsep}{3.5pt}
\renewcommand{\arraystretch}{0.90}
\begin{tabular}{llccccc}
\toprule
Noise & Method & CW-FID $\downarrow$ & CW-Cov. $\uparrow$ & CAS $\uparrow$ & FID $\downarrow$ & Cov $\uparrow$ \\
\midrule
Clean
& OT-FM & $11.42\scriptstyle\pm 0.19$ & $0.951\scriptstyle\pm 0.001$ & $0.938\scriptstyle\pm 0.002$ & $2.67\scriptstyle\pm 0.03$ & $0.974\scriptstyle\pm 0.002$ \\
\midrule
\multirow{6}{*}{Sym.\ 0.2}
& OT-FM & $19.23\scriptstyle\pm 0.20$ & $0.926\scriptstyle\pm 0.001$ & $0.745\scriptstyle\pm 0.002$ & $3.20\scriptstyle\pm 0.07$ & $0.968\scriptstyle\pm 0.000$ \\
& OT-FM (L1) & $21.86\scriptstyle\pm 0.16$ & $0.900\scriptstyle\pm 0.001$ & $0.743\scriptstyle\pm 0.004$ & $6.02\scriptstyle\pm 0.13$ & $0.953\scriptstyle\pm 0.007$ \\
& OT-FM (Huber) & $19.06\scriptstyle\pm 0.38$ & $0.927\scriptstyle\pm 0.002$ & $0.747\scriptstyle\pm 0.002$ & $3.13\scriptstyle\pm 0.22$ & $0.969\scriptstyle\pm 0.002$ \\
& FM-DivideMix & $19.10\scriptstyle\pm 0.03$ & $0.928\scriptstyle\pm 0.004$ & $0.747\scriptstyle\pm 0.001$ & $4.99\scriptstyle\pm 0.12$ & $0.968\scriptstyle\pm 0.002$ \\
\cmidrule(lr){2-7}
& FM-CCS, $\tau{=}0.001$ & $17.07\scriptstyle\pm 0.12$ & $0.927\scriptstyle\pm 0.001$ & {\bfseries\boldmath $0.944\scriptstyle\pm 0.002$} & $4.88\scriptstyle\pm 0.13$ & {\bfseries\boldmath $0.972\scriptstyle\pm 0.002$} \\
& FM-CCS, $\tau{=}0.01$ & {\bfseries\boldmath $15.26\scriptstyle\pm 0.08$} & {\bfseries\boldmath $0.940\scriptstyle\pm 0.001$} & $0.806\scriptstyle\pm 0.002$ & {\bfseries\boldmath $3.12\scriptstyle\pm 0.08$} & $0.971\scriptstyle\pm 0.001$ \\
\midrule
\multirow{6}{*}{Sym.\ 0.4}
& OT-FM & $33.10\scriptstyle\pm 0.23$ & $0.887\scriptstyle\pm 0.003$ & $0.547\scriptstyle\pm 0.000$ & $3.19\scriptstyle\pm 0.16$ & $0.968\scriptstyle\pm 0.003$ \\
& OT-FM (L1) & $35.26\scriptstyle\pm 0.19$ & $0.852\scriptstyle\pm 0.002$ & $0.549\scriptstyle\pm 0.003$ & $6.46\scriptstyle\pm 0.04$ & $0.951\scriptstyle\pm 0.006$ \\
& OT-FM (Huber) & $32.93\scriptstyle\pm 0.30$ & $0.887\scriptstyle\pm 0.006$ & $0.547\scriptstyle\pm 0.006$ & $3.16\scriptstyle\pm 0.06$ & $0.967\scriptstyle\pm 0.001$ \\
& FM-DivideMix & $31.20\scriptstyle\pm 0.79$ & $0.876\scriptstyle\pm 0.007$ & $0.551\scriptstyle\pm 0.008$ & $5.24\scriptstyle\pm 0.38$ & $0.957\scriptstyle\pm 0.002$ \\
\cmidrule(lr){2-7}
& FM-CCS, $\tau{=}0.001$ & {\bfseries\boldmath $16.90\scriptstyle\pm 0.03$} & {\bfseries\boldmath $0.925\scriptstyle\pm 0.000$} & {\bfseries\boldmath $0.894\scriptstyle\pm 0.005$} & $4.79\scriptstyle\pm 0.11$ & {\bfseries\boldmath $0.970\scriptstyle\pm 0.002$} \\
& FM-CCS, $\tau{=}0.01$ & $25.92\scriptstyle\pm 0.64$ & $0.910\scriptstyle\pm 0.002$ & $0.624\scriptstyle\pm 0.006$ & {\bfseries\boldmath $3.13\scriptstyle\pm 0.19$} & $0.970\scriptstyle\pm 0.001$ \\
\midrule
\multirow{6}{*}{Asym.\ 0.4}
& OT-FM & $16.29\scriptstyle\pm 0.22$ & $0.932\scriptstyle\pm 0.002$ & $0.795\scriptstyle\pm 0.002$ & $3.24\scriptstyle\pm 0.09$ & $0.968\scriptstyle\pm 0.001$ \\
& OT-FM (L1) & $19.90\scriptstyle\pm 0.18$ & $0.902\scriptstyle\pm 0.002$ & $0.786\scriptstyle\pm 0.004$ & $6.03\scriptstyle\pm 0.13$ & $0.951\scriptstyle\pm 0.003$ \\
& OT-FM (Huber) & $16.37\scriptstyle\pm 0.16$ & $0.931\scriptstyle\pm 0.002$ & $0.795\scriptstyle\pm 0.004$ & $3.24\scriptstyle\pm 0.04$ & $0.969\scriptstyle\pm 0.001$ \\
& FM-DivideMix & $16.24\scriptstyle\pm 0.16$ & $0.924\scriptstyle\pm 0.002$ & $0.794\scriptstyle\pm 0.002$ & $5.09\scriptstyle\pm 0.17$ & $0.966\scriptstyle\pm 0.004$ \\
\cmidrule(lr){2-7}
& FM-CCS, $\tau{=}0.001$ & $16.46\scriptstyle\pm 0.08$ & $0.925\scriptstyle\pm 0.002$ & {\bfseries\boldmath $0.928\scriptstyle\pm 0.003$} & $4.77\scriptstyle\pm 0.06$ & $0.970\scriptstyle\pm 0.003$ \\
& FM-CCS, $\tau{=}0.01$ & {\bfseries\boldmath $14.21\scriptstyle\pm 0.06$} & {\bfseries\boldmath $0.942\scriptstyle\pm 0.002$} & $0.843\scriptstyle\pm 0.001$ & {\bfseries\boldmath $3.17\scriptstyle\pm 0.07$} & {\bfseries\boldmath $0.970\scriptstyle\pm 0.004$} \\
\bottomrule
\end{tabular}%
\end{table}

\subsection{CIFAR-10 under label noise}
\label{subsec:exp_cifar_ynoise}

We evaluate symmetric label noise at rates $20\%$ and $40\%$ and asymmetric label noise at $40\%$. We use CW-FID, CW-Coverage, and CAS to assess class alignment. Table~\ref{tab:cifar10_ynoise_extended} shows substantial degradation under label noise. With $40\%$ symmetric noise, OT-FM's CW-FID increases from the clean reference of $11.42$ to $33.10$, while CAS falls from $0.938$ to $0.547$. FM-CCS substantially reduces this degradation: with $\tau=0.001$, CW-FID falls to $16.90$ and CAS rises to $0.894$. The improvements extend to the other noise settings. With $\tau=0.01$, FM-CCS reduces CW-FID from $19.23$ to $15.26$ under $20\%$ symmetric noise and from $16.29$ to $14.21$ under $40\%$ asymmetric noise.

The robust-loss and noisy-label baselines offer limited improvements over OT-FM, whereas FM-CCS achieves the lowest CW-FID across all three noise settings with the appropriate temperature. These results support the use of comparative residual scoring to improve class alignment. Figure~\ref{fig:cifar_main}(c)--(d) illustrates the same pattern: OT-FM generates plausible images that often do not match the requested class, while FM-CCS produces more consistent class-conditional samples. This improvement comes with a trade-off: at $\tau=0.001$, FM-CCS has higher unconditional FID than OT-FM despite its substantially better class alignment.

\subsection{Larger condition space: CIFAR-100}
\label{subsec:exp_cifar100}

To assess CCS with a larger class-conditional model, we evaluate CCS on CIFAR-100 under $40\%$ symmetric label noise, using $100$ classes and a model approximately $2.25$ times larger than the CIFAR-10 model. Both temperatures improve all three class-conditional metrics over OT-FM, with $1.44\times$--$1.48\times$ the runtime (Table~\ref{tab:cifar100_ynoise}). Appendix~\ref{app:large_condition_space} reports a complementary $1000$-class experiment.

\begin{table}[h!]
\centering
\caption{\textbf{CIFAR-100 under $40\%$ symmetric label noise.}
Values are mean $\pm$ standard deviation. Runtime is in minutes.}
\label{tab:cifar100_ynoise}
\scriptsize
\setlength{\tabcolsep}{3.5pt}
\renewcommand{\arraystretch}{0.95}
\begin{tabular}{lcccc}
\toprule
Method & CW-FID $\downarrow$ & CW-Cov. $\uparrow$
       & Classifier accuracy $\uparrow$ & Runtime $\downarrow$ \\
\midrule
OT-FM
& $100.33\pm0.54$
& $0.8272\pm0.0071$
& $0.3788\pm0.0032$
& $243.42\pm4.62$ \\
FM-CCS, $\tau_y=0.001$
& $\mathbf{83.16}\pm0.42$
& $\mathbf{0.8671}\pm0.0064$
& $\mathbf{0.5582}\pm0.0027$
& $350.57\pm4.34$ \\
FM-CCS, $\tau_y=0.01$
& $87.34\pm0.36$
& $0.8342\pm0.0058$
& $0.5231\pm0.0024$
& $360.36\pm3.62$ \\
\bottomrule
\end{tabular}
\end{table}

\subsection{Real-world noisy labels}
\label{subsec:exp_webvision}

\begin{wraptable}{r}{0.62\textwidth}
\vspace{-5mm}
\centering
\caption{\textbf{WebVision real-world noisy-label results.}
Models are trained on noisy WebVision labels and evaluated against clean ImageNet images restricted to the same 50 classes.}
\label{tab:webvision_results}
\small
\setlength{\tabcolsep}{3.2pt}
\renewcommand{\arraystretch}{0.92}
\begin{tabular}{lccc}
\toprule
Method & CAS $\uparrow$ & CW-FID $\downarrow$ & CW-Cov. $\uparrow$ \\
\midrule
OT-FM
& $0.3456{\scriptstyle\pm 0.0030}$
& $65.11{\scriptstyle\pm 0.26}$
& $0.7743{\scriptstyle\pm 0.0040}$ \\
FM-DivideMix
& $0.3535{\scriptstyle\pm 0.0055}$
& $65.09{\scriptstyle\pm 0.61}$
& $0.7825{\scriptstyle\pm 0.0106}$ \\
FM-CCS
& {\bfseries\boldmath $0.5414{\scriptstyle\pm 0.0049}$}
& {\bfseries\boldmath $60.13{\scriptstyle\pm 0.50}$}
& {\bfseries\boldmath $0.8074{\scriptstyle\pm 0.0030}$} \\
\bottomrule
\end{tabular}
\vspace{-4mm}
\end{wraptable}

We evaluate FM-CCS on the standard 50-class Mini WebVision setting~\citep{li2017webvision,jiang2018mentornet}, using naturally noisy web labels for training and clean ImageNet images~\citep{russakovsky2015imagenet} from the same classes for evaluation. Table~\ref{tab:webvision_results} shows that FM-CCS improves CAS, CW-FID, and CW-Coverage over both OT-FM and FM-DivideMix. Implementation details are given in Appendices~\ref{app:hyperparameters} and~\ref{app:corruptions}.

\section{Conclusion}

We studied robust FM under target corruption and label noise without access to clean pairs or explicit noise models. The two corruptions affect VF regression differently: target corruption perturbs the interpolation state and velocity target, whereas label noise changes the conditional VF evaluated on the same trajectory. This motivates TRS and CCS, which convert temporal and comparative residual evidence into sample weights. Experiments on 2D benchmarks, CIFAR-10, CIFAR-100, and WebVision demonstrate improved robustness, with additional evidence under more challenging label noise and simultaneous target and label corruption.

\paragraph{Limitations.}
TRS requires informative residual separation and may downweight rare valid structures or retain easily fitted corruptions. Its late-biased sampling also changes the temporal weighting of optimization, whose effect may depend on the dataset, model, and probability path. CCS requires distinguishable conditional losses, is developed for discrete labels, and has refresh cost that grows with the number of conditions. Our joint-corruption evaluation is limited to controlled 2D settings. Broader corruption models, severe class imbalance, and matched comparisons with diffusion-based methods remain directions for future work. Appendix~\ref{app:limitations} discusses these limitations in detail.

\begin{ack}
Computational resources were provided on the Berzelius system
funded by the Knut and Alice Wallenberg Foundation and operated by NAISS.

Mert Can Kurucu was supported by the Scientific and Technological
Research Council of T{\"u}rkiye (T{\"U}B{\.I}TAK) through the
2219 International Postdoctoral Research Fellowship Programme.

Erik Englesson was supported by the Wallenberg AI, Autonomous Systems and Software Program~(WASP) funded by the Knut and Alice Wallenberg Foundation. 

Hossein Azizpour acknowledges the funding from Knut and Alice Wallenberg Foundation, KTH Digital Futures, and the Swedish e-Science Research Centre and the support by the Wallenberg AI, Autonomous Systems and Software Program~(WASP).

The authors declare no competing interests.
\end{ack}

\bibliographystyle{unsrt}
\bibliography{references}

\appendix
\newpage

\section{Derivations and Theoretical Supporting Results}
\label{app:proofs}

We derive Propositions~\ref{prop:x_noise_decomposition} and~\ref{prop:y_noise_decomposition} directly from the definitions of the FM path,
target velocity, and residual under the two noise corruptions.

\subsection{Derivation of Proposition~1}
\label{app:derivation_prop1}

Let the noise $\Delta_1$ be defined as,
\[
\Delta_1 := \tilde{x}_1 - x_1,
\qquad\text{so that}\qquad
\tilde{x}_1 = x_1 + \Delta_1.
\]
Under the standard linear FM path,
\[
x_t = (1-t)x_0 + t x_1,
\qquad
u(x_0,x_1)=x_1-x_0,
\]
while the corrupted interpolation path and corrupted velocity target are
\[
\tilde{x}_t = (1-t)x_0 + t\tilde{x}_1,
\qquad
\tilde{u}(x_0,\tilde{x}_1)=\tilde{x}_1-x_0.
\]

Substituting $\tilde{x}_1=x_1+\Delta_1$ gives
\begin{align*}
\tilde{x}_t
&= (1-t)x_0 + t(x_1+\Delta_1) \\
&= \bigl((1-t)x_0 + t x_1\bigr) + t\Delta_1 \\
&= x_t + t\Delta_1,
\end{align*}
and
\begin{align*}
\tilde{u}(x_0,\tilde{x}_1)
&= (x_1+\Delta_1)-x_0 \\
&= (x_1-x_0)+\Delta_1 \\
&= u(x_0,x_1)+\Delta_1.
\end{align*}

The noisy-target residual is therefore
\begin{align*}
r_\theta^x(x_0,t;\tilde{x}_1,y)
&:= v_{\theta,t}(\tilde{x}_t,y)-\tilde{u}(x_0,\tilde{x}_1) \\
&= v_{\theta,t}(x_t+t\Delta_1,y)-\bigl(u(x_0,x_1)+\Delta_1\bigr).
\end{align*}
Adding and subtracting $v_{\theta,t}(x_t,y)$,
\begin{align*}
r_\theta^x(x_0,t;\tilde{x}_1,y)
&= \bigl(v_{\theta,t}(x_t,y)-u(x_0,x_1)\bigr) \\
&\quad + \bigl(v_{\theta,t}(x_t+t\Delta_1,y)-v_{\theta,t}(x_t,y)\bigr)
-\Delta_1.
\end{align*}
With
\[
r_\theta^{\mathrm{clean}}(x_0,t;x_1,y)
:= v_{\theta,t}(x_t,y)-u(x_0,x_1),
\]
we obtain
\[
r_\theta^x
=
r_\theta^{\mathrm{clean}}
+
\bigl(v_{\theta,t}(x_t+t\Delta_1,y)-v_{\theta,t}(x_t,y)\bigr)
-\Delta_1.
\]

This is the decomposition in Proposition~1. The corruption enters through a
time-independent target-shift term $-\Delta_1$ and a state-perturbation term
evaluated at the displaced point $x_t+t\Delta_1$. Since
\[
\|\tilde{x}_t-x_t\| = \|t\Delta_1\| = t\|\Delta_1\|,
\]
the geometric effect of the corrupted state grows linearly with $t$ along the
path.

\subsection{Derivation of Proposition~2}
\label{app:derivation_prop2}

Under label noise, the target sample $x_1$ is unchanged and only the
observed condition is corrupted from $y$ to $\tilde{y}$. Hence the path and
target velocity remain
\[
x_t=(1-t)x_0+t x_1,
\qquad
u(x_0,x_1)=x_1-x_0.
\]
The noisy-condition residual is
\[
r_\theta^y(x_0,t;x_1,\tilde{y})
:= v_{\theta,t}(x_t,\tilde{y})-u(x_0,x_1).
\]
Adding and subtracting $v_{\theta,t}(x_t,y)$,
\begin{align*}
r_\theta^y(x_0,t;x_1,\tilde{y})
&= \bigl(v_{\theta,t}(x_t,y)-u(x_0,x_1)\bigr)
 + \bigl(v_{\theta,t}(x_t,\tilde{y})-v_{\theta,t}(x_t,y)\bigr) \\
&= r_\theta^{\mathrm{clean}}(x_0,t;x_1,y)
 + \Delta_{\theta,t}(x_t; y \to \tilde{y}),
\end{align*}
where
\[
\Delta_{\theta,t}(x_t; y \to \tilde{y})
:= v_{\theta,t}(x_t,\tilde{y})-v_{\theta,t}(x_t,y).
\]

This is the decomposition in Proposition~2. In contrast to the target corruption case, there is no perturbed interpolation state and no shifted target velocity. The
entire corruption effect is concentrated in the conditional vector-field
difference at the same clean state $x_t$, which is precisely the structure
used by CCS.

The following result formalizes the sign convention used by CCS under a class-conditional margin assumption.

\begin{proposition}[CCS sign separation under class-conditional margin]
\label{prop:ccs_sign}
Fix a clean pair $(x_0, x_1)$ with clean label $y$ and a probe-time set $\mathcal{T}_y \subset [0,1]$. Assume there exist pointwise margins $\{\gamma_t\}_{t \in \mathcal{T}_y}$ with $\gamma_t > 0$ such that, for every $t \in \mathcal{T}_y$ and every competing label $c \neq y$,
\begin{equation}
\ell_\theta^{y}(x_0, t, x_1; y) + \gamma_t \;\le\; \ell_\theta^{y}(x_0, t, x_1; c),
\label{eq:app_margin_assumption}
\end{equation}
and let $\bar{\gamma} := \frac{1}{|\mathcal{T}_y|} \sum_{t \in \mathcal{T}_y} \gamma_t > 0$. Then the aggregated CCS score $M_i$ defined in~\eqref{eq:method_y_ccs} satisfies:
\begin{enumerate}
    \item[(i)] If the observed label is clean, $\tilde{y} = y$, then $M_i \le -\bar{\gamma}$.
    \item[(ii)] If the observed label is corrupted, $\tilde{y} \neq y$, then $M_i \ge \bar{\gamma}$.
\end{enumerate}
In particular, clean and corrupted observed labels produce CCS margins with opposite signs, separated by at least $2\bar{\gamma}$.
\end{proposition}

\begin{proof}
\textbf{(i) Clean case, $\tilde{y} = y$.} By~\eqref{eq:app_margin_assumption}, for every competitor $c \neq y$ and every $t \in \mathcal{T}_y$,
\[
\ell_\theta^{y}(x_0, t, x_1; y) \le \ell_\theta^{y}(x_0, t, x_1; c) - \gamma_t.
\]
Taking the minimum over $c \neq y$ on the right-hand side,
\[
\ell_\theta^{y}(x_0, t, x_1; y) \le \min_{c \neq y} \ell_\theta^{y}(x_0, t, x_1; c) - \gamma_t,
\]
so that, under the CCS convention
$s_\theta^{y}=\ell_\theta^{y}(\tilde y)-\min_{c\neq \tilde y}\ell_\theta^{y}(c)$, the instantaneous margin satisfies
\[
s_\theta^{y}(x_0, t, x_1, y)
=
\ell_\theta^{y}(x_0, t, x_1; y)
-
\min_{c \neq y} \ell_\theta^{y}(x_0, t, x_1; c)
\le -\gamma_t.
\]
Averaging over $\mathcal{T}_y$ gives $M_i \le -\bar{\gamma}$.

\textbf{(ii) Corrupted case, $\tilde{y} \neq y$.} By~\eqref{eq:app_margin_assumption} applied to the corrupted label $\tilde{y}$, viewed as a competitor of the clean label $y$,
\[
\ell_\theta^{y}(x_0, t, x_1; y) + \gamma_t \le \ell_\theta^{y}(x_0, t, x_1; \tilde{y}).
\]
Since $\tilde{y} \neq y$, the clean label $y$ is one element of the competitor set $\{c : c \neq \tilde{y}\}$, so
\[
\min_{c \neq \tilde{y}} \ell_\theta^{y}(x_0, t, x_1; c) \le \ell_\theta^{y}(x_0, t, x_1; y).
\]
Combining the two inequalities,
\[
s_\theta^{y}(x_0, t, x_1, \tilde{y})
=
\ell_\theta^{y}(x_0, t, x_1; \tilde{y})
-
\min_{c \neq \tilde{y}} \ell_\theta^{y}(x_0, t, x_1; c)
\ge
\ell_\theta^{y}(x_0, t, x_1; \tilde{y})
-
\ell_\theta^{y}(x_0, t, x_1; y)
\ge \gamma_t.
\]
Averaging over $\mathcal{T}_y$ gives $M_i \ge \bar{\gamma}$.
\end{proof}

\begin{remark}
Proposition~\ref{prop:ccs_sign} gives a sufficient condition under which zero is an appropriate decision center for the CCS weight map $w_i^y = \sigma(-M_i / \tau_y)$: once the class-conditional vector fields have separated enough that~\eqref{eq:app_margin_assumption} holds on average, clean and corrupted observed labels fall on opposite sides of zero with a guaranteed gap of $2\bar{\gamma}$. Clean labels have negative CCS scores and therefore receive near-unit weights, while corrupted labels have positive CCS scores and are downweighted. The sigmoid softly interpolates between these two regimes, with $\tau_y$ controlling the sharpness of the transition. This is also why CCS is not applied from the start of training: the margin assumption fails early, when the class-conditional fields have not yet separated.
\end{remark}

\section{Implementation Details}
\label{app:implementation}

\subsection{Experimental protocol}
\label{app:protocol}

We evaluate target corruption and label noise on 2D benchmarks and CIFAR-10, and additionally evaluate label noise on CIFAR-100 and WebVision. In the \emph{target corruption} setting, the observed data sample is corrupted. In the \emph{label-noise} setting, the data sample remains clean but its conditioning label is corrupted. The 2D and CIFAR-10 experiments use controlled corruptions to isolate the two mechanisms, CIFAR-100 extends the controlled label-noise evaluation to a larger class-conditional setting, and WebVision tests the label-noise method under naturally occurring web-label noise. For each benchmark and corruption type, all compared methods use the same source distribution, network architecture, optimizer, batch size, training budget, and sampling solver. Only the robustness mechanism is changed. All results are reported as mean $\pm$ standard deviation over three random seeds.

\subsection{Architectures and optimization}
\label{app:hyperparameters}

\paragraph{2D benchmarks.}
For unconditional experiments under target corruption, we use a $4$-layer MLP with hidden width $256$, SiLU activations, and sinusoidal time embeddings of dimension $64$ \citep{lipman2023flow}. For class-conditional experiments under label noise, we use a $3$-layer MLP with the same hidden width and learned class embeddings of dimension $32$. All 2D models are trained with Adam, learning rate $10^{-3}$, and batch size $4096$. Sampling uses Euler integration with $20$ steps. For evaluation, we draw $50{,}000$ samples in the unconditional setting and $5{,}000$ samples per class in the class-conditional setting.

\paragraph{CIFAR-10.}
For CIFAR-10, we use the standard FM UNet from the official Flow Matching codebase~\citep{lipman2023flow}: base width $128$, channel multipliers $[2,2,2]$, four residual blocks per resolution, self-attention at $16\times16$, and dropout $0.3$. The same backbone, optimizer, probability path, and sampler are used for both target corruption and label-noise experiments; class conditioning is used in the label-noise experiments. We train for $1000$ epochs with AdamW, learning rate $10^{-4}$, $\beta_1=0.9$, $\beta_2=0.95$, and batch size $64$. We maintain an exponential moving average of model weights and sample with the Heun solver using $100$ function evaluations.

\paragraph{CIFAR-100.}
We use a conditional UNet with base width $192$, channel multipliers $[2,2,2]$, four residual blocks per resolution, attention at downsampling factor $2$, dropout $0.3$, and conditioning over $100$ classes. This model has $125{,}273{,}091$ parameters, compared with $55{,}682{,}051$ for the conditional CIFAR-10 model. We train under $40\%$ symmetric label noise for $600$ epochs using AdamW with learning rate $10^{-4}$, $\beta_1=0.9$, $\beta_2=0.95$, and batch size $64$ per GPU across eight GPUs, giving a global batch size of $512$. The learning rate remains constant, and the model-weight EMA decay is capped at $0.999$.

For this experiment, CCS uses a single probe time, $\mathcal{T}_y=\{0.5\}$, with one source draw per sample and comparisons against all $100$ labels. We perform one full-dataset refresh halfway through training and retain the resulting weights for the remaining updates. We evaluate temperatures $\tau_y\in\{0.001,0.01\}$.

\paragraph{WebVision.}
For WebVision, we use the same UNet backbone as in the CIFAR-10 label-noise experiments, with class conditioning over $50$ classes. All images are resized to $32\times32$. OT-FM, FM-DivideMix, and FM-CCS are trained for $300$ epochs with AdamW, learning rate $10^{-4}$, batch size $128$, EMA weights, and Heun sampling with $100$ function evaluations.

\subsection{Corruption models}
\label{app:corruptions}

We use controlled corruption models on 2D benchmarks, CIFAR-10, and CIFAR-100 to isolate the two failure modes studied in the main paper. In target corruption experiments, only a minority of samples is corrupted, matching the robust-regression setting where strong outliers are present but the clean distribution remains identifiable. In label-noise experiments, the target sample remains clean but the observed label is corrupted according to standard discrete noisy-label protocols \citep{na2024labelnoise}. WebVision is used only for naturally occurring real-world label noise; no synthetic target corruption is added.

\paragraph{2D target corruption.}
For the 2D target corruption setting, we add corrupted samples equal to $20\%$ of the clean training-set size. These samples are drawn from an off-manifold Gaussian blob,
$\mathcal{N}(\mu_{\mathrm{blob}},\,0.3^2 I)$,
where $\mu_{\mathrm{blob}}$ is placed at Euclidean distance $4.0$ from the clean data centroid. This creates a compact corrupted minority that is visually separated from the clean manifold. The purpose of this setting is not to model all possible corruptions, but to make the outlier geometry explicit and to test whether the method suppresses the influence of corrupted targets without distorting the clean distribution.

\paragraph{2D label noise.}
For label noise, we use segmented spirals with $C=10$ classes and consider three standard regimes: symmetric flips at rates $20\%$ and $40\%$, and asymmetric flips at rate $40\%$. Under symmetric noise with rate $\eta$, the observed label is kept unchanged with probability $1-\eta$ and otherwise flipped uniformly among the other $C-1$ classes:
\[
\mathbb{P}(\tilde y = y \mid y)=1-\eta,
\qquad
\mathbb{P}(\tilde y = c \mid y)=\frac{\eta}{C-1},\quad c\neq y.
\]
Under asymmetric noise, labels are flipped according to an adjacent-class permutation, so the corrupted label remains another valid class rather than arbitrary noise, following standard noisy-label protocols~\citep{patrini2017making}. We additionally provide qualitative results on a two-class Moons dataset in Appendix~\ref{app:additional_samples}; in the two-class case, asymmetric and symmetric noise are equivalent.

\paragraph{CIFAR-10 target corruption.}
For CIFAR-10 target corruption, we corrupt $20\%$ of the training images. This keeps corrupted samples as a minority of strong outliers while preserving the clean distribution as the dominant signal. In the \textbf{Gaussian} setting, we add pixel noise with standard deviation $\sigma=0.3$ in $[0,1]$. In the \textbf{SP} setting, $15\%$ of the pixels in each corrupted image are independently set to either $0$ or $1$~\citep{kurucu2025fractional}. Both corruptions are severe at the sample level, so they clearly degrade standard OT-FM, but the corruption rate is low enough that recovering the clean distribution remains meaningful.

\paragraph{CIFAR-10 label noise.}
For CIFAR-10 label noise, we use symmetric label flips at rates $20\%$ and $40\%$, and asymmetric flips at rate $40\%$, following the standard class-dependent noisy-label protocol of~\citep{patrini2017making}. In the asymmetric setting, semantically related classes are mapped into one another:
\[
\texttt{truck}\rightarrow\texttt{automobile},\quad
\texttt{bird}\rightarrow\texttt{airplane},\quad
\texttt{cat}\leftrightarrow\texttt{dog},\quad
\texttt{deer}\rightarrow\texttt{horse},
\]
while the remaining classes are left unchanged. This creates structured label noise in which the corrupted label is often plausible, making conditional alignment more challenging than in purely random relabeling.

\paragraph{CIFAR-100 label noise.}
For CIFAR-100, we use $40\%$ symmetric label noise, following the same symmetric-flip protocol as above: each selected label is replaced uniformly by one of the other $99$ classes.

\paragraph{WebVision real-world label noise.}
For the real-world noisy-label experiment, we use the $50$-class Mini WebVision setting~\citep{jiang2018mentornet}, consisting of the first $50$ ILSVRC-2012 synsets from the Google image subset of WebVision~\citep{li2017webvision}. The training set contains $65{,}944$ web images with naturally noisy labels. No synthetic target corruption or synthetic label flipping is added.

\subsection{Method-specific settings}
\label{app:our_methods}

\paragraph{FM-TRS.}
FM-TRS is used for target corruption experiments on 2D benchmarks and CIFAR-10. For 2D target corruption, the default configuration uses EMA decay $\beta=0.99$, quantile centering at the $70$th percentile, and activates reweighting after the first $25\%$ of training. We report temperatures $\tau\in\{0.001,0.01\}$ to show the effect of the score-to-weight sharpness. The \emph{uniform} variant samples FM times uniformly, $t\sim\mathcal{U}(0,1)$, while the default \emph{late-biased} variant uses the time-sampling rule described in Section~\ref{subsec:method_x}. For CIFAR-10 target corruption, we use the same temperature settings and compare the same two time-sampling variants. The EMA decay is $\beta=0.999$, and reweighting is activated after the first $25\%$ of training. Ablations for warmup, EMA decay, quantile centering, and late-biased time sampling are given in Appendix~\ref{app:ablations}.

\paragraph{FM-CCS.}
FM-CCS is used for label-noise experiments on 2D benchmarks, CIFAR-10, CIFAR-100, and WebVision. For 2D, CIFAR-10, and WebVision, we refresh the comparative score buffer at $25\%$ and $50\%$ of training and use the fixed probe set $\mathcal{T}_y=\{0.2,\,0.5,\,0.8\}.$ For 2D and CIFAR-10, we report temperatures $\tau_y\in\{0.001,0.01\}$, while WebVision uses $\tau_y=0.001$. CIFAR-100 uses a single probe at $t=0.5$ and one full-dataset refresh halfway through training, with $\tau_y\in\{0.001,0.01\}$. Reweighting starts at the first refresh, and no relabeling is performed; CCS acts only through sample reweighting. Ablations on the probe set and refresh schedule are given in Appendix~\ref{app:ablations}.

\subsection{Adapted Baselines}
\label{app:adapted_baselines}

All baselines are implemented inside the same FM training pipeline as the proposed methods. OT-FM, OT-FM (L1), and OT-FM (Huber) are used as generic regression baselines where applicable. FM-DeepGUM is used only for target corruption experiments, while FM-DivideMix is used only for label-noise experiments, including WebVision.

\paragraph{OT-FM, OT-FM (L1), and OT-FM (Huber).}
For each corruption setting, the baselines use the same observed FM residual as the corresponding noisy objective: $r_\theta^x$ in Eq.~\eqref{eq:x_noise_residual_main} for target corruption and $r_\theta^y$ in Eq.~\eqref{eq:y_noise_residual} for label noise. Standard OT-FM minimizes the squared residual,
\[
\ell_i^{\mathrm{MSE}}=\|r_{\theta,i}\|_2^2.
\]
OT-FM (L1) replaces this pointwise penalty with
\[
\ell_i^{\mathrm{L1}}=\|r_{\theta,i}\|_1.
\]
OT-FM (Huber) applies the Huber penalty elementwise with threshold $\delta=1.0$:
\[
\ell_i^{\mathrm{Huber}}
=
\sum_{j=1}^{d}
h_\delta\!\bigl(r_{\theta,i,j}\bigr),
\qquad
h_\delta(z)=
\begin{cases}
\frac{1}{2}z^2, & |z|\le \delta,\\[4pt]
\delta\left(|z|-\frac{1}{2}\delta\right), & |z|>\delta.
\end{cases}
\]
These baselines modify only the pointwise regression penalty; they do not introduce score buffers, sample selection, time-dependent reliability weighting, or comparisons across alternative class-conditional VFs.

\paragraph{FM-DeepGUM.}
We adapt DeepGUM~\citep{lathuiliere2018deepgum} only to the target corruption experiments. The original method addresses robust regression by treating the per-sample squared error as evidence for a Gaussian--uniform mixture: an inlier Gaussian component models clean residuals, a uniform component models gross outliers, and the posterior probability of the inlier component is used as a soft weight on the regression loss.

To port DeepGUM to FM, we apply this mixture model to the per-sample squared FM residual under the target corruption objective in Eq.~\eqref{eq:x_noise_objective}. We maintain an exponential moving average of this residual score for each training sample, fit a Gaussian to the lower $(1-\pi_{\mathrm{out}})$ quantile of the population EMA scores, and use the resulting inlier posterior as a soft weight on the FM objective. For stability and efficiency in the FM setting, we make two implementation choices: \emph{(i)} we replace the per-iteration EM updates of the Gaussian and mixture parameters by a one-shot robust estimate of the inlier Gaussian from the lower-quantile mass, and \emph{(ii)} we replace the bounded uniform component, whose support has no natural choice for an unbounded loss-valued score, by a flat constant proportional to the estimated inlier scale. The FM backbone, probability path, optimizer, and sampler are otherwise unchanged. FM-DeepGUM therefore provides a generic robust-regression baseline that does not exploit the FM-specific time-dependent residual structure.
\paragraph{FM-DivideMix.}
We adapt DivideMix~\citep{li2020dividemix} only to the label-noise experiments. Original DivideMix paper combines co-divide, co-refinement, co-guessing, and MixMatch. The original DivideMix ablation shows that this co-divide mechanism alone already improves substantially over the warm-up baseline at high noise rates, while MixMatch and the pseudo-labeling components provide additional classification-specific gains~\citep{li2020dividemix}. We therefore use FM-DivideMix as a co-divide-style adaptation of DivideMix to FM, because MixMatch, co-refinement, or co-guessing components rely on softmax label predictions and soft pseudo-labels, whereas a class-conditional FM network outputs a velocity field conditioned on the observed label. Constructing an analogous label posterior for FM would require either an additional classifier or evaluating the velocity field across candidate labels together with a new pseudo-labeling rule. We therefore retain the observed labels for all samples and use the GMM-based peer weights as the label-noise adaptation. The FM architecture, probability path, optimizer, and sampler are otherwise unchanged.

Two FM networks $A$ and $B$ with parameters $\theta_A$ and $\theta_B$ are co-trained. After a short warm-up phase of standard FM training on the observed noisy labels, we periodically recompute, for each network independently, a per-sample loss score under the observed label,
\[
L_i^{(N)}
=
\frac{1}{T}
\sum_{m=1}^{T}
\ell_{\theta_N}^{y}(x_{0,i}^{(m)},t_i^{(m)},x_{1,i};\tilde y_i),
\qquad N\in\{A,B\},
\]
where $x_{0,i}^{(m)}\sim p_0$ and $t_i^{(m)}\sim\mathcal{U}(0,1)$. A two-component Gaussian mixture is fit to the population $\{L_i^{(N)}\}_{i=1}^{|\mathcal{D}|}$, and the posterior probability of the lower-mean component is used as the per-sample clean probability $p_i^{(N)}\in[0,1]$.

To reduce self-confirmation bias, the clean probabilities from one network weight the FM loss of the peer network. For a mini-batch $\mathcal{B}$, network $B$ is updated with
\[
\mathcal{L}_{\mathrm{FM\text{-}DivideMix}}^{(B)}(\theta_B)
=
\frac{1}{\sum_{i\in\mathcal{B}} p_i^{(A)}}
\sum_{i\in\mathcal{B}}
p_i^{(A)}
\ell_{\theta_B}^{y}(x_{0,i},t_i,x_{1,i};\tilde y_i),
\]
and $\theta_A$ is updated symmetrically using $p_i^{(B)}$. Thus, samples assigned high clean probability by the peer network contribute more strongly, while high-loss samples are softly downweighted.

\subsection{Metrics}
\label{app:metrics}

For 2D experiments, evaluation references are drawn from the same clean data-generating distribution used to define the benchmark. For CIFAR-10, we evaluate against the standard clean test split. For WebVision, we evaluate against clean ImageNet images restricted to the same $50$ classes as the Mini WebVision setting \citep{na2024labelnoise}. All metrics are first computed per seed and then averaged across seeds.

\paragraph{2D target corruption.}
We report $W_2$, MMD, and Coverage using $50{,}000$ generated samples. $W_2$ measures global transport discrepancy, and MMD provides a complementary kernel-based distribution discrepancy. Coverage follows the precision/recall-style manifold-aware evaluation framework of \citep{naeem2020reliable}.

\paragraph{2D label noise.}
Under label noise, the main failure mode is cross-class mixing, so we report class-wise metrics. For any class-wise metric $M_c$, we compute
\[
\mathrm{CW}\text{-}M=\frac{1}{C}\sum_{c=1}^{C} M_c.
\]
We report CW-$W_2$ and CW-Coverage using $5{,}000$ generated samples per class.

\paragraph{Joint corruption and the 1000-class experiment.}
For these additional experiments, we compute the empirical Wasserstein-2 distance separately for each class using squared Euclidean transport costs and uniform sample weights. We take the square root of the optimal transport cost for each class and then average the resulting distances equally across classes, reporting CW-$W_2$. The experiment-specific sample counts are given in Appendices~\ref{app:joint_corruption} and~\ref{app:large_condition_space}.

\paragraph{CIFAR-10 target corruption.}
We report FID, Inception Score (IS), and Coverage on $50{,}000$ generated samples. FID is the main sample-quality metric~\citep{heusel2017gans}; IS captures sharpness and class diversity~\citep{salimans2016improved}; and Coverage measures support recovery relative to the clean reference distribution~\citep{naeem2020reliable}. We additionally report a clean oracle trained on uncorrupted CIFAR-10 under the same backbone and optimization budget.

\paragraph{CIFAR-10 label noise.}
For label noise, unconditional realism alone is not sufficient: a model may generate plausible CIFAR-like images while failing to respect the requested class. We therefore report FID together with class-wise CW-FID, CW-Coverage, and the Classification Accuracy Score (CAS)~\citep{ravuri2019cas}. CW-FID is computed per class against the corresponding clean class distribution and averaged across classes. CAS measures how well the class signal is preserved in the generated distribution by training a classifier on generated samples and evaluating it on real test data.

\paragraph{CIFAR-100 label noise.}
We report CW-FID, CW-Coverage, and classifier accuracy. Classifier accuracy is the fraction of generated images assigned the requested label by a fixed ResNet-18 trained on real CIFAR-100 images.

\paragraph{WebVision real-world label noise.}
For WebVision, we report CAS, CW-FID, and CW-Coverage. CAS is computed over $50{,}000$ generated samples, with $1{,}000$ samples per class. CW-FID and CW-Coverage are computed class-wise against clean ImageNet reference images from the same $50$ classes and then averaged across classes. These metrics match the class-conditional failure mode targeted by CCS under naturally noisy labels.

\section{Ablations}
\label{app:ablations}

We ablate the main design choices behind FM-TRS and FM-CCS on 2D benchmarks. The goal is not exhaustive hyperparameter tuning, but to verify whether the design choices suggested by the analysis are empirically meaningful. For FM-TRS, we study warmup, EMA decay, quantile-based weighting, and whether late-biased time sampling alone explains the gains. For FM-CCS, we study how often the comparative score needs to be refreshed. Overall, the results support the main design choices: TRS benefits from stable residual aggregation and late-time scoring, while CCS does not require frequent score recomputation. Ablation results are obtained from independent runs with different seeds.

\paragraph{When should TRS reweighting start?}
Table~\ref{tab:abl_trs_warmup} varies the warmup fraction before activating TRS reweighting. Starting at $10\%$, $25\%$, or $50\%$ gives competitive results, with $25\%$ yielding the best mean $W_2$ and MMD among the tested settings. We use this as a practical default. Warmup allows the VF to begin fitting the data before residual scores determine sample weights, but its appropriate duration may depend on the dataset, architecture, and optimization dynamics.

\begin{table}[h!]
\centering
\caption{\textbf{TRS warmup fraction} (2D target corruption). Mean $\pm$ std over 3 seeds.}
\label{tab:abl_trs_warmup}
\small
\setlength{\tabcolsep}{6pt}
\begin{tabular}{lccc}
\toprule
Warmup & $W_2 \downarrow$ & MMD $\downarrow$ & Coverage $\uparrow$ \\
\midrule
10\% & $0.463\scriptstyle\pm 0.015$ & $0.054\scriptstyle\pm 0.007$ & $0.896\scriptstyle\pm 0.003$ \\
25\% & {\bfseries\boldmath $0.412\scriptstyle\pm 0.048$} &  {\bfseries\boldmath $0.045\scriptstyle\pm 0.009$} & {\bfseries\boldmath $0.897\scriptstyle\pm 0.002$} \\
50\% & $0.442\scriptstyle\pm 0.034$  & $0.049\scriptstyle\pm 0.012$ & $0.895\scriptstyle\pm 0.002$ \\
\bottomrule
\end{tabular}
\end{table}

\paragraph{How sensitive is TRS to the EMA decay?}
Table~\ref{tab:abl_trs_ema} varies the EMA decay $\beta$ used to accumulate TRS scores. The tested values produce $W_2$ values of $0.551$, $0.434$, and $0.736$ for $\beta=0.9$, $0.99$, and $0.999$, respectively. Thus, EMA decay has a meaningful effect, although all three settings improve over the standard OT-FM reference. The results are consistent with a trade-off between responsiveness and smoothing: smaller values give greater influence to recent stochastic residual evaluations, whereas larger values retain older scores for longer. We use $\beta=0.99$ for the 2D experiments, while some tuning may help across architectures and training setups.

\begin{table}[h!]
\centering
\caption{\textbf{TRS EMA decay} (2D target corruption). Mean $\pm$ std over 3 seeds.}
\label{tab:abl_trs_ema}
\small
\setlength{\tabcolsep}{5pt}
\begin{tabular}{lccc}
\toprule
$\beta$ & $W_2 \downarrow$ & MMD $\downarrow$ & Coverage $\uparrow$ \\
\midrule
0.9 & $0.551\scriptstyle\pm 0.081$ & $0.079\scriptstyle\pm 0.015$ & {\bfseries\boldmath $0.897\scriptstyle\pm 0.003$} \\
0.99 & {\bfseries\boldmath $0.434\scriptstyle\pm 0.033$} & {\bfseries\boldmath $0.048\scriptstyle\pm 0.013$} & {\bfseries\boldmath $0.897\scriptstyle\pm 0.002$} \\
0.999 & $0.736\scriptstyle\pm 0.046$ & $0.140\scriptstyle\pm 0.018$ & $0.892\scriptstyle\pm 0.002$ \\
\bottomrule
\end{tabular}
\end{table}

\paragraph{How sensitive is TRS to the quantile hyperparameter?}
Table~\ref{tab:abl_trs_quantile} varies the score-buffer quantile used as the transition center in the TRS weight map. This quantile is a robustness hyperparameter, not a known corruption rate, and is selected without using corruption indicators. The $70$th percentile gives the best transport metrics among the tested quantiles, however it is not sensitive around 70th percentile, different values give similar performance. We use the $70$th percentile as the default, corresponding to $\rho_x=0.3$ in the TRS quantile rule.

\begin{table}[h!]

\centering

\caption{\textbf{TRS quantile hyperparameter} (2D target corruption). The percentile denotes the score-buffer quantile used as the transition center, not a known or estimated corruption rate. Mean $\pm$ std over 3 seeds.}

\label{tab:abl_trs_quantile}

\small

\setlength{\tabcolsep}{5pt}

\begin{tabular}{lccc}

\toprule

Quantile & $W_2 \downarrow$ & MMD $\downarrow$ & Coverage $\uparrow$ \\

\midrule

50th & $0.564\scriptstyle\pm 0.037$ & $0.061\scriptstyle\pm 0.014$ & $0.889\scriptstyle\pm 0.002$ \\

70th & {\bfseries\boldmath $0.431\scriptstyle\pm 0.079$} & {\bfseries\boldmath $0.047\scriptstyle\pm 0.014$} & {\bfseries\boldmath $0.896\scriptstyle\pm 0.002$}  \\

90th & $0.443\scriptstyle\pm 0.036$ & $0.049\scriptstyle\pm 0.011$ & $0.894\scriptstyle\pm 0.001$\\

\bottomrule

\end{tabular}

\end{table}

\paragraph{Is late-biased time sampling sufficient by itself?}
Table~\ref{tab:abl_trs_timebias_only} isolates late-biased FM time sampling from TRS reweighting. Late-biased sampling alone improves over OT-FM, reducing $W_2$ from $1.208$ to $0.970$ and MMD from $0.281$ to $0.179$. However, the full FM-TRS method is substantially stronger, reaching $W_2=0.441$ and MMD $=0.049$. Relative to OT-FM, late-biased sampling alone improves $W_2$ by about $20\%$ and MMD by about $36\%$, whereas FM-TRS improves them by about $64\%$ and $83\%$. Thus, late-time exposure is slightly useful, but the main gain comes from using this residual information for sample reweighting.

\begin{table}[h!]
\centering
\caption{\textbf{Ablation of late-biased time sampling without reweighting} on 2D target corruption. OT-FM with late-biased sampling alone improves only modestly over OT-FM, while FM-TRS yields substantially larger gains, showing that the benefit of TRS does not come from time bias alone.}
\label{tab:abl_trs_timebias_only}
\small
\setlength{\tabcolsep}{5pt}
\begin{tabular}{lccc}
\toprule
Method & $W_2 \downarrow$ & MMD $\downarrow$ & Coverage $\uparrow$ \\
\midrule
OT-FM & $1.208\scriptstyle\pm 0.099$ & $0.281\scriptstyle\pm 0.033$ & $0.855\scriptstyle\pm 0.018$ \\
OT-FM (Late-biased sampling) & $0.970\scriptstyle\pm 0.141$ & $0.179\scriptstyle\pm 0.037$ & $0.864\scriptstyle\pm 0.002$ \\
FM-TRS, $\tau{=}0.01$ & {\bfseries\boldmath $0.441\scriptstyle\pm 0.031$} & {\bfseries\boldmath $0.049\scriptstyle\pm 0.010$} & {\bfseries\boldmath $0.895\scriptstyle\pm 0.001$} \\
\bottomrule
\end{tabular}
\end{table}

\paragraph{How often should CCS scores be refreshed?}

Table~\ref{tab:abl_ccr_refresh} varies the timing of the first CCS refresh and the number of subsequent refreshes under $40\%$ symmetric label noise on the 2D benchmark. The first refresh also ends warmup. Starting at either $25\%$ or $50\%$ of training remains effective, indicating that performance does not depend critically on one exact refresh epoch in this experiment. We use refreshes at $25\%$ and $50\%$ because this schedule gives the best mean result while requiring only two dataset-wide comparative computations.

\begin{table}[h!]

\centering

\caption{\textbf{CCS refresh schedule} (2D $40\%$ symmetric label noise). Mean $\pm$ std over 3 seeds.}

\label{tab:abl_ccr_refresh}

\small

\setlength{\tabcolsep}{5pt}

\begin{tabular}{lcc}

\toprule

Refresh points & CW-$W_2 \downarrow$ & CW-Cov. $\uparrow$ \\

\midrule

25\% & $0.221\scriptstyle\pm 0.004$ & $0.726\scriptstyle\pm 0.026$ \\

25\%, 50\% & {\bfseries\boldmath $0.201\scriptstyle\pm 0.008$} & {\bfseries\boldmath $0.734\scriptstyle\pm 0.031$} \\

25\%, 50\%, 75\% & $0.228\scriptstyle\pm 0.009$ & $0.724\scriptstyle\pm 0.022$ \\

50\% & $0.230\scriptstyle\pm 0.015$ & $0.718\scriptstyle\pm 0.016$ \\

\bottomrule

\end{tabular}

\end{table}

\subsection{Time sampling on clean data}
\label{app:clean_time_sampling}

To isolate the effect of late-biased optimization, we compare OT-FM trained with uniform times against OT-FM trained with $\pi_x(t)=2t$, without sample reweighting. All other training and evaluation settings are identical within each comparison. Tables~\ref{tab:clean_time_2d} and~\ref{tab:clean_time_cifar10} report results on clean 2D data and clean CIFAR-10, respectively.

\begin{table}[h!]
\centering
\caption{\textbf{Time sampling on clean 2D data.}
Neither method uses sample reweighting.
Values are mean $\pm$ standard deviation.}
\label{tab:clean_time_2d}
\small
\setlength{\tabcolsep}{5pt}
\begin{tabular}{llccc}
\toprule
Dataset & Time sampling & $W_2\downarrow$
        & MMD $\downarrow$ & Coverage $\uparrow$ \\
\midrule
2 Spirals & $\mathcal{U}(0,1)$
& $0.5618\pm0.0109$
& $0.0955\pm0.0045$
& $0.8830\pm0.0019$ \\
2 Spirals & $\pi_x(t)=2t$
& $0.4843\pm0.0467$
& $0.0898\pm0.0126$
& $0.8879\pm0.0001$ \\
\midrule
Moons & $\mathcal{U}(0,1)$
& $0.3977\pm0.0088$
& $0.1268\pm0.0038$
& $0.8806\pm0.0003$ \\
Moons & $\pi_x(t)=2t$
& $0.3583\pm0.0150$
& $0.1106\pm0.0092$
& $0.8816\pm0.0004$ \\
\bottomrule
\end{tabular}
\end{table}

\begin{table}[h!]
\centering
\caption{\textbf{Time sampling on clean CIFAR-10.}
Neither method uses sample reweighting.
Mean $\pm$ standard deviation over three seeds.}
\label{tab:clean_time_cifar10}
\small
\setlength{\tabcolsep}{7pt}
\begin{tabular}{lccc}
\toprule
Time sampling & FID $\downarrow$
              & IS $\uparrow$ & Coverage $\uparrow$ \\
\midrule
$\mathcal{U}(0,1)$
& $2.67\pm0.03$
& $9.53\pm0.06$
& $0.971\pm0.002$ \\
$\pi_x(t)=2t$
& $2.62\pm0.04$
& $9.59\pm0.03$
& $0.973\pm0.005$ \\
\bottomrule
\end{tabular}
\end{table}

Late-biased optimization does not degrade the reported mean metrics in these clean-data experiments. Together with Table~\ref{tab:abl_trs_timebias_only}, these results distinguish the effect of temporal weighting from the robustness gains obtained through TRS. They support the implementation choice in the evaluated settings, while its effect may depend on the dataset, source distribution, model, and optimization settings.

\subsection{CCS probe-time selection}
\label{app:ccs_probe_times}

We vary the size and composition of $\mathcal{T}_y$ on the 2D benchmark under $40\%$ symmetric label noise. Table~\ref{tab:abl_ccs_probes} shows that a single midpoint probe is less effective than probe sets spanning different regions of the trajectory. Two separated probes recover most of the improvement, while $\{0.2,0.5,0.8\}$ gives the best mean CW-$W_2$ among the tested sets. Increasing the set to five probes provides no further improvement in this experiment. We therefore use three probes as a practical balance between performance and scoring cost.

\begin{table}[h!]
\centering
\caption{\textbf{CCS probe-time ablation}
under $40\%$ symmetric label noise on the 2D benchmark.
Values are mean $\pm$ standard deviation.}
\label{tab:abl_ccs_probes}
\small
\setlength{\tabcolsep}{7pt}
\begin{tabular}{lcc}
\toprule
$\mathcal{T}_y$ & CW-$W_2\downarrow$ & CW-Cov. $\uparrow$ \\
\midrule
$\{0.5\}$
& $0.3634\pm0.0387$ & $0.6917\pm0.0123$ \\
$\{0.2,0.8\}$
& $0.1874\pm0.0030$ & $0.7090\pm0.0107$ \\
$\{0.2,0.5,0.8\}$
& $\mathbf{0.1774}\pm0.0088$
& $\mathbf{0.7143}\pm0.0244$ \\
$\{0.2,0.35,0.5,0.65,0.8\}$
& $0.1944\pm0.0100$ & $0.7136\pm0.0142$ \\
\bottomrule
\end{tabular}
\end{table}

\section{Additional Experimental Analyses}
\label{app:additional_experimental_analyses}

\subsection{More challenging label noise}
\label{app:complex_label_noise}

We extend the segmented-spirals evaluation to $60\%$ symmetric label noise and $40\%$ instance-dependent label noise. In the instance-dependent setting, samples near class boundaries are more likely to have their labels flipped, with an expected overall noise rate of $40\%$. Clean labels are used only to construct the synthetic corruption and are not provided to CCS during training.

\begin{table}[h!]
\centering
\caption{\textbf{More challenging label noise on segmented spirals.}
Mean $\pm$ standard deviation over three seeds.}
\label{tab:complex_label_noise}
\small
\setlength{\tabcolsep}{5pt}
\begin{tabular}{llcc}
\toprule
Noise & Method & CW-$W_2\downarrow$ & CW-Cov. $\uparrow$ \\
\midrule
\multirow{3}{*}{$60\%$ symmetric}
& OT-FM
& $1.7778\pm0.0265$ & $0.4416\pm0.0269$ \\
& FM-CCS, $\tau_y=0.001$
& $0.2924\pm0.0362$ & $0.6822\pm0.0785$ \\
& FM-CCS, $\tau_y=0.01$
& $0.2907\pm0.0364$ & $0.6911\pm0.0712$ \\
\midrule
\multirow{3}{*}{$40\%$ instance-dependent}
& OT-FM
& $1.1848\pm0.0138$ & $0.5944\pm0.0103$ \\
& FM-CCS, $\tau_y=0.001$
& $0.9798\pm0.0174$ & $0.6225\pm0.0214$ \\
& FM-CCS, $\tau_y=0.01$
& $0.9874\pm0.0171$ & $0.6179\pm0.0245$ \\
\bottomrule
\end{tabular}
\end{table}

Both CCS temperatures improve CW-$W_2$ and CW-Coverage in both settings. The best mean CW-$W_2$ decreases by approximately $84\%$ under $60\%$ symmetric noise and $17\%$ under instance-dependent noise. These results complement the asymmetric-noise and naturally noisy WebVision evaluations in the main paper.

\subsection{Simultaneous target corruption and label noise}
\label{app:joint_corruption}

\paragraph{Experimental setup.}
We construct a training pool of $50{,}000$ points from the 10-class segmented-spirals distribution. We first apply symmetric label noise at rate $20\%$ or $40\%$, replacing each selected label uniformly with a different class. We then replace the coordinates of $10{,}000$ points with samples from $\mathcal{N}((4,0),0.3^2 I_2)$, retaining their observed labels. Thus, target outliers constitute $20\%$ of the final dataset, and both inliers and outliers are eligible for label corruption.

We use the same conditional MLP, optimizer, and batch size as in the 2D label-noise experiments described in Appendix~\ref{app:hyperparameters}. Training uses $10{,}001$ updates and source distribution $\mathcal{N}(0,4I_2)$. OT-FM uses uniform training times, whereas the combined method uses $t=\sqrt{U}$ with $U\sim\mathcal{U}(0,1)$.

For the combined method, TRS uses EMA decay $0.99$ and the $70$th percentile of observed TRS scores as the weighting threshold. Both weight maps use temperature $0.01$. CCS uses probe times $\{0.2,0.5,0.8\}$ and refreshes at steps $2500$ and $5000$. Training uses unit weights before the first refresh and the product of the TRS and CCS weights afterward. The training loss is the minibatch mean of the weighted residual losses, without renormalizing the sample weights.

For evaluation, we generate $5000$ points in total, with $500$ points per class, using $20$ Euler steps. The clean reference pool contains $10{,}000$ points.

\begin{table}[h!]
\centering
\caption{\textbf{Simultaneous target and label corruption on segmented spirals.}
Mean $\pm$ standard deviation over three seeds.}
\label{tab:joint_corruption}
\small
\setlength{\tabcolsep}{7pt}
\begin{tabular}{lcc}
\toprule
Corruption setting & OT-FM CW-$W_2\downarrow$
                   & FM-TRS+CCS CW-$W_2\downarrow$ \\
\midrule
$20\%$ target outliers + $20\%$ label flips
& $1.857\pm0.052$ & $0.218\pm0.009$ \\
$20\%$ target outliers + $40\%$ label flips
& $2.079\pm0.063$ & $0.286\pm0.022$ \\
\bottomrule
\end{tabular}
\end{table}

The combined weighting reduces CW-$W_2$ by approximately $88\%$ and $86\%$, respectively, providing initial evidence that the two corruption-specific weights can be combined in this controlled setting. More complex datasets motivate further study of score interactions, relative calibration, and alternative combination rules.

\subsection{A controlled experiment with 1000 classes}
\label{app:large_condition_space}

We evaluate CCS on a 1000-class Gaussian grid under $40\%$ symmetric label noise to isolate the effect of a large condition space from the additional complexity of image generation.

\paragraph{Experimental setup.}
We define $1000$ Gaussian classes on a two-dimensional $40\times25$ grid, with means spanning $[-6,6]\times[-3.75,3.75]$ and covariance $0.0025I_2$. The training pool contains $50{,}000$ points with uniformly sampled class labels, giving $50$ points per class in expectation.

We use the same conditional MLP and training configuration as in Appendix~\ref{app:joint_corruption}, with the class embedding extended to $1000$ labels. Both OT-FM and FM-CCS use uniform training times. CCS compares all classes using temperature $0.01$, probe times $\{0.2,0.5,0.8\}$, and refreshes at $25\%$ and $50\%$ of training. The first comparative margin is used directly, and subsequent margins are smoothed with EMA decay $0.99$.

For evaluation, we generate $200{,}000$ points, with exactly $200$ samples per class, using $20$ Euler steps. The clean reference pool also contains $200{,}000$ points.

\begin{table}[h!]
\centering
\caption{\textbf{A 1000-class Gaussian-grid experiment}
under $40\%$ symmetric label noise.}
\label{tab:1000_class}
\small
\setlength{\tabcolsep}{10pt}
\begin{tabular}{lcc}
\toprule
Method & CW-$W_2\downarrow$ & Runtime (s) $\downarrow$ \\
\midrule
OT-FM & $3.395\pm0.143$ & $17.4$ \\
FM-CCS & $0.078\pm0.002$ & $21.2$ \\
\bottomrule
\end{tabular}
\end{table}

CCS substantially reduces CW-$W_2$ while requiring approximately $1.22\times$ the baseline training time, providing a controlled test of CCS with a large condition space.

\subsection{Additional density evaluation}
\label{app:density}

We additionally report Density alongside Coverage for the 2D target corruption experiment. Table~\ref{tab:additional_density} shows the same qualitative trend as the $W_2$ and MMD comparisons: FM-TRS improves both reported metrics relative to OT-FM.

\begin{table}[h!]
\centering
\caption{\textbf{Additional density evaluation under 2D target corruption.}
Values are mean $\pm$ standard deviation.}
\label{tab:additional_density}
\small
\setlength{\tabcolsep}{10pt}
\begin{tabular}{lcc}
\toprule
Method & Density $\uparrow$ & Coverage $\uparrow$ \\
\midrule
OT-FM & $138.45\pm1.13$ & $0.861\pm0.018$ \\
FM-TRS & $159.95\pm1.37$ & $0.897\pm0.002$ \\
\bottomrule
\end{tabular}
\end{table}

\subsection{Computational cost}
\label{app:computational_cost}

We report computational overhead using measured wall-clock training time over the full 1000-epoch CIFAR-10 runs. This gives a practical comparison because our methods use the same FM backbone, optimizer, training budget, and sampling setup, and differ only in the robustness mechanism.

Table~\ref{tab:cost_main} shows that FM-TRS adds negligible overhead relative to OT-FM. Standard OT-FM takes approximately 16h 08m, while FM-TRS takes 16h 15m, corresponding to a relative cost of $1.007\times$. Thus, the score accumulation and sample reweighting in TRS add less than $1\%$ overhead over the full run.

FM-CCS is slightly more expensive. However, with the sparse two-refresh schedule used in our experiments, the total training time is 17h 19m, corresponding to $1.072\times$ the cost of OT-FM. This is a modest overhead of roughly $7\%$, and it does not require changing the FM backbone or training pipeline.

The CIFAR-10 and Mini WebVision experiments were run on NVIDIA A100 GPUs with 40\,GB memory, using a single GPU per training run.

\begin{table}[h!]
\centering
\caption{\textbf{Measured training-time overhead on CIFAR-10 over the full 1000-epoch run.} Relative cost is reported with respect to standard OT-FM under the same backbone and optimization setup.}
\label{tab:cost_main}
\small
\setlength{\tabcolsep}{8pt}
\begin{tabular}{lcc}
\toprule
Method & Training time & Relative cost \\
\midrule
OT-FM & $\sim$16h 08m & $1.000\times$ \\
FM-TRS & $\sim$16h 15m & $1.007\times$ \\
FM-CCS & $\sim$17h 19m & $1.072\times$ \\
\bottomrule
\end{tabular}
\end{table}

\paragraph{Scaling with the condition space.}
An exact CCS refresh requires $O(N|\mathcal{T}_y|C)$ sample--condition evaluations, where $C=|\mathcal{Y}|$. The comparisons are vectorized and performed only at scheduled refresh points, rather than during every optimization step. Nevertheless, their practical cost increases with the number of conditions, dataset size, and model-evaluation complexity.

The measured relative runtime is approximately $1.22\times$ on the controlled 1000-class Gaussian grid (Table~\ref{tab:1000_class}) and $1.44\times$-$1.48\times$ on CIFAR-100 (Table~\ref{tab:cifar100_ynoise}). These ratios compare CCS with OT-FM within the corresponding experimental setup. Larger-scale applications may benefit from approximate comparisons using sampled or top-$k$ alternatives, hierarchical screening, or partial refreshes. These are directions for future work rather than components of the evaluated method.

\subsection{Statistical significance}
\label{app:statistical_significance}

We additionally performed paired two-sided $t$-tests across matched random seeds for the main comparisons. For 2D target corruption experiments on shifted spirals, FM-TRS significantly improves the primary distributional metrics over OT-FM, OT-FM (L1), OT-FM (Huber), and FM-DeepGUM. The improvements in $W_2$ and MMD are significant in all these comparisons ($p \le 0.009$). The late-time TRS variant also significantly improves over the uniform-time TRS variant on $W_2$ ($p=0.018$) and MMD ($p=0.033$), supporting the role of time-biased scoring. For 2D label-noise experiments, FM-CCS significantly reduces CW-$W_2$ across all symmetric and asymmetric noise settings against OT-FM, OT-FM (L1), OT-FM (Huber), and FM-DivideMix ($p \le 0.004$). 

On CIFAR-10 with Gaussian target corruption, FM-TRS significantly improves FID, IS, and Coverage over OT-FM, OT-FM (L1), OT-FM (Huber), and FM-DeepGUM ($p \le 0.008$). Compared with the uniform-time TRS variant, the late-time variant significantly improves both FID and IS ($p=0.011$). On CIFAR-10 label noise, FM-CCS significantly improves the primary class-conditional metrics: CW-FID is significantly reduced for all noise settings against OT-FM, OT-FM (L1), OT-FM (Huber), and FM-DivideMix ($p \le 0.002$), and CAS is also significantly improved in all corresponding comparisons ($p \le 0.001$).

On WebVision, FM-CCS also significantly improves over OT-FM on all reported
class-conditional metrics: CAS ($p=5.49\times 10^{-4}$), CW-FID ($p=0.00742$),
and CW-Coverage ($p=0.00745$). These results support the main conclusion that the proposed scores improve robustness consistently across seeds, especially on the primary metrics targeted by each corruption setting. We report these aggregate statistics only as a consistency check, and avoid over-interpreting marginal $p$-values due to the small number of seeds.

\section{Score Diagnostics}
\label{app:diagnostics}

This section visualizes the score behavior underlying FM-TRS and FM-CCS. These diagnostics are not meant to evaluate final generation quality; instead, they test whether the proposed scores capture the reliability signals predicted by the analysis. For both methods, we use the same convention: larger scores indicate less reliable samples and therefore lead to smaller weights. For target corruption, this signal is the time-dependent growth of residual separation along the FM path. For label noise, it is whether an alternative class-conditional VF explains the same clean trajectory better than the observed class-conditional VF.

\begin{figure}[h!]
    \centering
    \includegraphics[width=\linewidth]{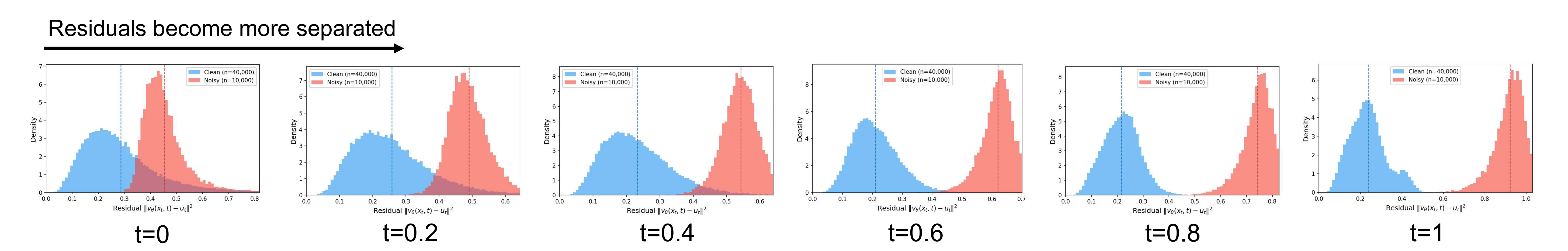}
    \caption{\textbf{TRS score behavior across FM time.} Residual-based separation between clean and corrupted samples becomes stronger at intermediate-to-late FM times. This supports the FM-TRS design: under target corruption, later points on the interpolation path expose the geometric effect of endpoint corruption more clearly, making residual-based reliability scoring more informative than at early times.}
    \label{fig:trs_time_behavior}
\end{figure}

Figure~\ref{fig:trs_time_behavior} shows the residual signal used by FM-TRS across FM time on CIFAR-10. Consistent with Section~\ref{subsec:x_noise_in_fm}, clean and corrupted samples are harder to distinguish near the start of the path, but become increasingly separable at intermediate-to-late times as the corrupted interpolation state moves away from its clean counterpart. This supports the late-biased score construction used by FM-TRS.

\begin{figure}[h!]
    \centering
    \includegraphics[width=\linewidth]{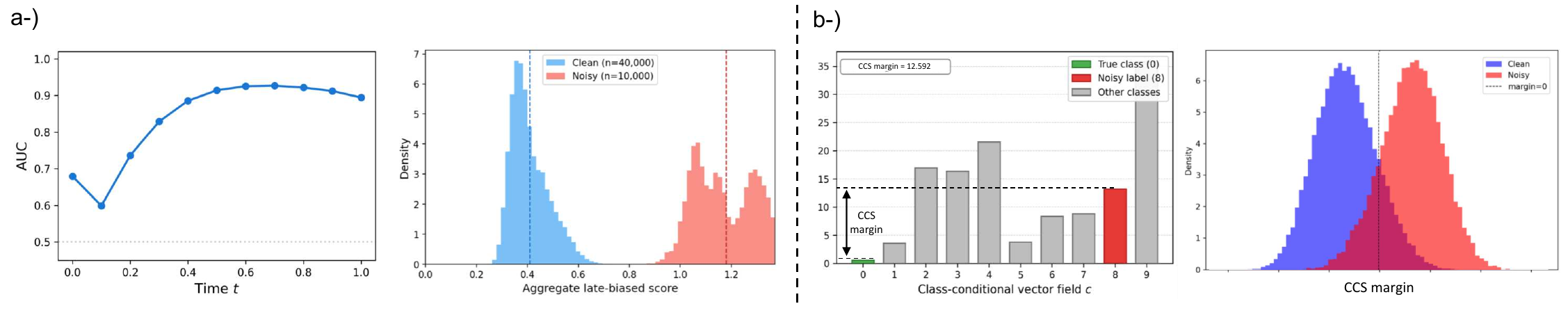}
    \caption{\textbf{Aggregated score separation under target corruption and label noise.} \textbf{Left:} FM-TRS score distribution under target corruption. Aggregated residual scores separate clean and corrupted samples, with corrupted samples concentrated in the high-score region that receives lower weights. \textbf{Right:} FM-CCS score distribution under label noise. The comparative margin separates clean and label-flipped samples around zero: positive scores indicate that an alternative class-conditional VF explains the trajectory better than the observed label, while negative scores indicate that the observed label explains the trajectory better than the alternatives.}
    \label{fig:score_behavior}
\end{figure}

Figure~\ref{fig:score_behavior} shows the aggregated score distributions for both methods. Under target corruption, FM-TRS separates clean and corrupted samples after residual aggregation, with corrupted samples concentrated in the high-score region targeted by downweighting. Under label noise, FM-CCS separates clean and label-flipped samples around the zero-centered margin: positive scores indicate that some alternative class-conditional VF explains the trajectory better than the observed label, and therefore provide evidence of unreliable label supervision; negative scores indicate that the observed label explains the trajectory better than the alternatives.

\newpage
\section{Pseudocode}
\label{app:pseudocode}

Algorithms~\ref{alg:x_noise} and~\ref{alg:y_noise} summarize the training procedures for FM-TRS and FM-CCS.

\begin{algorithm}
\caption{Robust FM under target corruption (FM-TRS)}
\label{alg:x_noise}
\begin{algorithmic}[1]
\Require Corrupted set $\{(\tilde{x}_{1,i}, y_i)\}_{i=1}^N$, source $p_0$, time distribution $\pi_x$, EMA coefficient $\beta_x$, tail fraction $\rho_x$, temperature $\tau_x$, warmup step $K_{\mathrm{warm}}$, training steps $K$
\State Initialize $\theta$ and score buffer $S_i \gets 0$ for all $i$
\For{$k = 1, \dots, K$}
    \State Sample minibatch $\mathcal{B} \subset \{1, \dots, N\}$
    \For{$i \in \mathcal{B}$}
        \State Sample $x_{0,i} \sim p_0$, $t_i \sim \pi_x$
        \State $\tilde{x}_{t,i} \gets (1-t_i)x_{0,i}+t_i\tilde{x}_{1,i}$, \quad $\tilde{u}_i \gets \tilde{x}_{1,i}-x_{0,i}$
        \State $\ell_i \gets \|v_{\theta,t_i}(\tilde{x}_{t,i},y_i)-\tilde{u}_i\|_2^2$
        \State $S_i \gets \beta_x S_i + (1-\beta_x)\ell_i$
    \EndFor
    \If{$k < K_{\mathrm{warm}}$}
        \State $w_i \gets 1$ for all $i \in \mathcal{B}$
    \Else
        \State $q_x \gets Q_{1-\rho_x}(\{S_j\}_{j=1}^N)$
        \State $w_i \gets \sigma\!\left(-(S_i-q_x)/\tau_x\right)$ for all $i \in \mathcal{B}$ \Comment{stop-gradient}
    \EndIf
    \State $\mathcal{L}_{\mathcal{B}} \gets \frac{1}{|\mathcal{B}|}\sum_{i \in \mathcal{B}} w_i\, \ell_i$
    \State Update $\theta$ via a gradient step on $\mathcal{L}_{\mathcal{B}}$
\EndFor
\end{algorithmic}
\end{algorithm}

\begin{algorithm}[h!]
\caption{Robust FM under label noise (FM-CCS)}
\label{alg:y_noise}
\begin{algorithmic}[1]
\Require Noisy-label set $\{(x_{1,i}, \tilde{y}_i)\}_{i=1}^N$, source $p_0$, time distribution $\pi_y$, probe set $\mathcal{T}_y$, refresh epochs $\mathcal{E}_{\mathrm{ref}}$, temperature $\tau_y$, total epochs $E$
\State Initialize $\theta$ and weights $w_i \gets 1$ for all $i$
\For{$e = 1, \dots, E$}
    \If{$e \in \mathcal{E}_{\mathrm{ref}}$}
        \For{$i = 1, \dots, N$}
            \State Sample $x_{0,i} \sim p_0$, set $u_i \gets x_{1,i}-x_{0,i}$ and $M_i \gets 0$
            \For{$t \in \mathcal{T}_y$}
                \State $x_{t,i} \gets (1-t)x_{0,i}+t x_{1,i}$
                \State $\ell_{\tilde y} \gets \|v_{\theta,t}(x_{t,i},\tilde{y}_i)-u_i\|_2^2$
                \State $\ell_{\min} \gets \min_{c \neq \tilde{y}_i}\|v_{\theta,t}(x_{t,i},c)-u_i\|_2^2$
                \State $M_i \gets M_i + (\ell_{\tilde y}-\ell_{\min})/|\mathcal{T}_y|$
            \EndFor
        \EndFor
        \State $w_i \gets \sigma(-M_i/\tau_y)$ for all $i$ \Comment{stop-gradient}
    \EndIf
    \For{minibatch $\mathcal{B}$}
        \For{$i \in \mathcal{B}$}
            \State Sample $x_{0,i} \sim p_0$, $t_i \sim \pi_y$
            \State $x_{t,i} \gets (1-t_i)x_{0,i}+t_i x_{1,i}$, \quad $u_i \gets x_{1,i}-x_{0,i}$
            \State $\ell_i \gets \|v_{\theta,t_i}(x_{t,i},\tilde{y}_i)-u_i\|_2^2$
        \EndFor
        \State $\mathcal{L}_{\mathcal{B}} \gets \frac{1}{|\mathcal{B}|}\sum_{i \in \mathcal{B}} w_i\, \ell_i$
        \State Update $\theta$ via a gradient step on $\mathcal{L}_{\mathcal{B}}$
    \EndFor
\EndFor
\end{algorithmic}
\end{algorithm}

\section{Additional Qualitative Samples}
\label{app:additional_samples}

This section provides complementary qualitative evidence for the main empirical findings. We show representative samples on the 2D benchmarks and on CIFAR-10 under both noise corruptions. The qualitative patterns are consistent with the quantitative results: under target corruption, FM-TRS suppresses corrupted off-manifold structure and better recovers the clean geometry or image manifold; under label noise, FM-CCS improves class consistency by reducing updates induced by incorrect observed conditions.

\subsection{2D qualitative results}
\label{app:qual_2d}

\begin{figure}[h]
    \centering
    \begin{subfigure}[t]{0.32\linewidth}
        \centering
        \includegraphics[width=\linewidth]{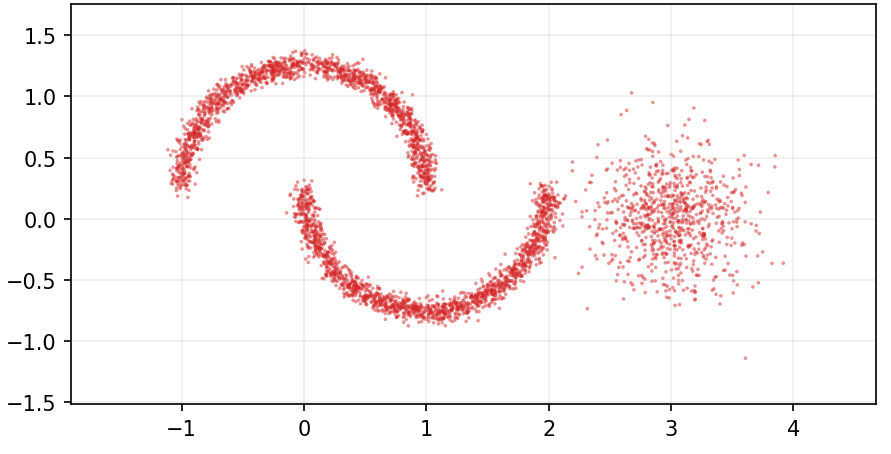}
        \caption{Corrupted training distribution}
        \label{fig:app_moons_xnoise_a}
    \end{subfigure}\hfill
    \begin{subfigure}[t]{0.32\linewidth}
        \centering
        \includegraphics[width=\linewidth]{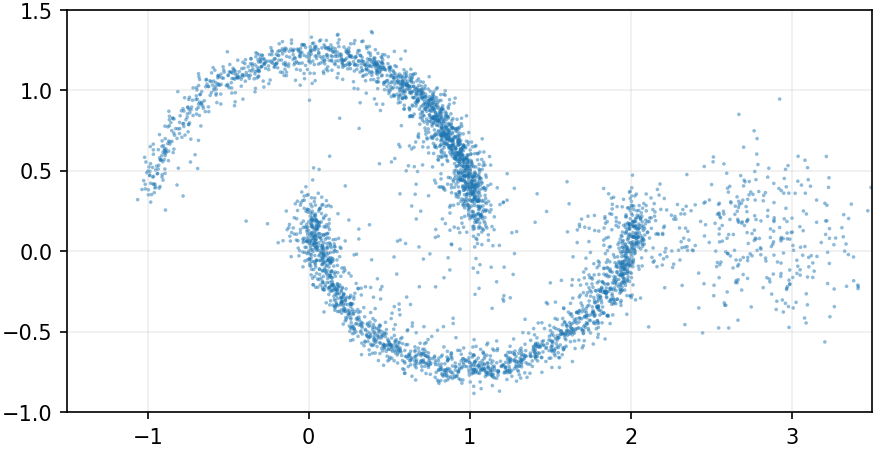}
        \caption{OT-FM}
        \label{fig:app_moons_xnoise_b}
    \end{subfigure}\hfill
    \begin{subfigure}[t]{0.32\linewidth}
        \centering
        \includegraphics[width=\linewidth]{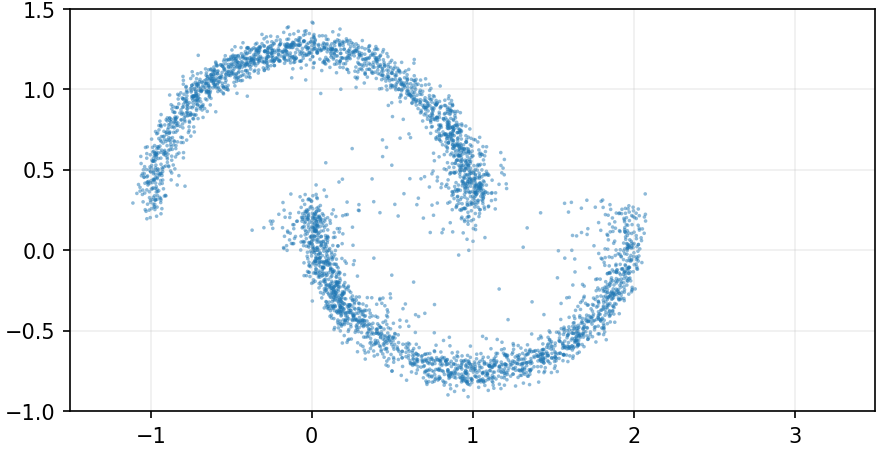}
        \caption{FM-TRS}
        \label{fig:app_moons_xnoise_c}
    \end{subfigure}
    \caption{\textbf{Additional 2D target corruption results on moons.} The corrupted training data contain an off-manifold outlier mode. Standard OT-FM partially reproduces this corrupted structure, whereas FM-TRS suppresses it and better recovers the clean moons geometry.}
    \label{fig:app_moons_xnoise}
\end{figure}

\begin{figure}[h]
    \centering
    \begin{subfigure}[t]{0.32\linewidth}
        \centering
        \includegraphics[width=\linewidth]{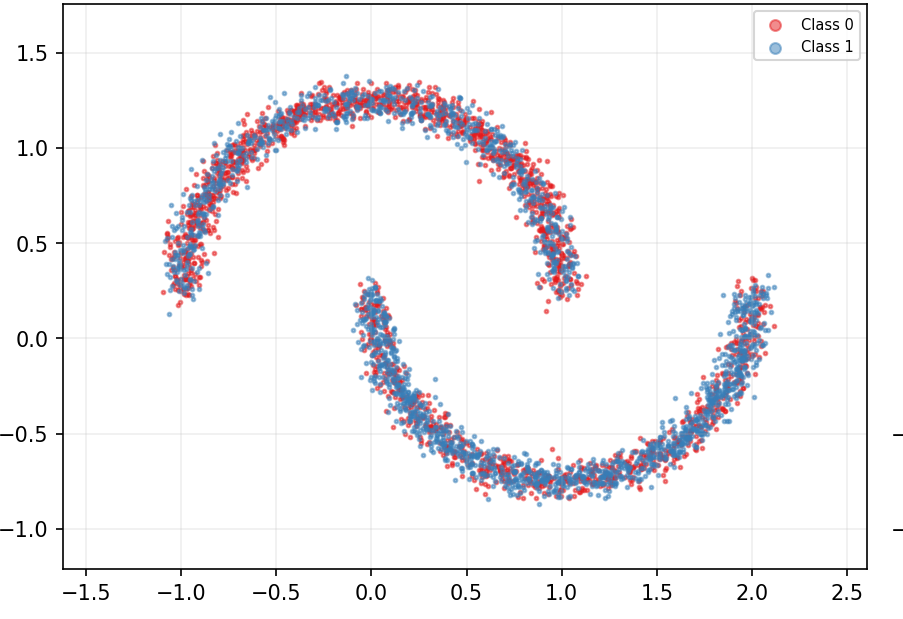}
        \caption{Training data with label noise}
        \label{fig:app_spirals_ynoise_a}
    \end{subfigure}\hfill
    \begin{subfigure}[t]{0.32\linewidth}
        \centering
        \includegraphics[width=\linewidth]{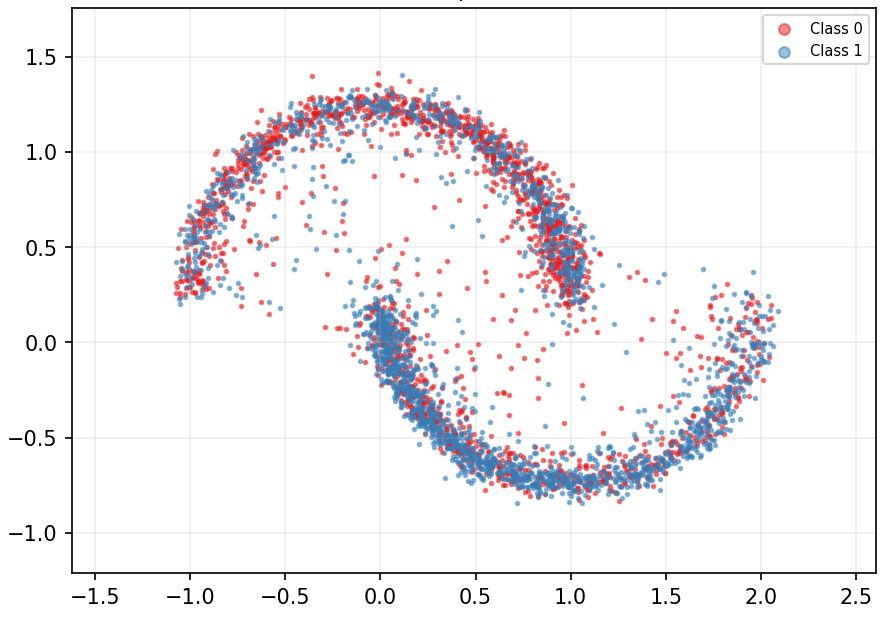}
        \caption{OT-FM}
        \label{fig:app_spirals_ynoise_b}
    \end{subfigure}\hfill
    \begin{subfigure}[t]{0.32\linewidth}
        \centering
        \includegraphics[width=\linewidth]{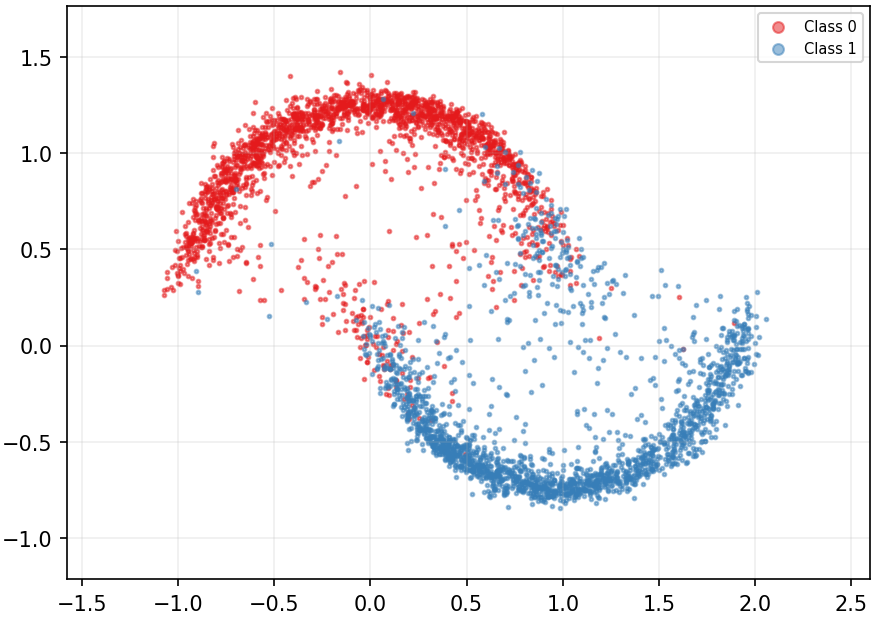}
        \caption{FM-CCS}
        \label{fig:app_spirals_ynoise_c}
    \end{subfigure}
    \caption{\textbf{Additional 2D label-noise results on segmented spirals.} Under noisy observed conditions, standard OT-FM shows substantial cross-class mixing. FM-CCS produces visibly cleaner class separation and better class-aligned spiral structure.}
    \label{fig:app_spirals_ynoise_scatter}
\end{figure}

Figures~\ref{fig:app_moons_xnoise} and~\ref{fig:app_spirals_ynoise_scatter} show additional qualitative results on the 2D moons and segmented spirals benchmarks.

Figure~\ref{fig:app_moons_xnoise} illustrates target corruption on the moons dataset. The corrupted training set contains a compact off-manifold minority mode. Standard OT-FM partially follows this corrupted structure and generates samples away from the clean moons manifold; moreover, the corrupted targets also distort the overall geometry of the generated moons. In contrast, FM-TRS suppresses the off-manifold mode and better preserves the clean data geometry.

Figure~\ref{fig:app_spirals_ynoise_scatter} illustrates label noise on segmented spirals. In this setting, the main failure mode is cross-class mixing rather than off-manifold generation. Standard OT-FM produces visibly entangled class regions, whereas FM-CCS yields cleaner class separation and better class-aligned spiral structure.

\subsection{CIFAR-10 qualitative results}
\label{app:qual_cifar}

\begin{figure}[t!]
    \centering
    \begin{subfigure}[t]{0.49\linewidth}
        \centering
        \includegraphics[width=\linewidth]{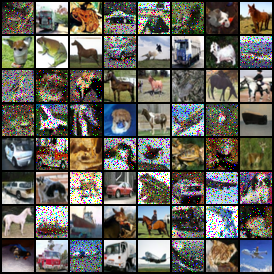}
        \caption{OT-FM}
        \label{fig:app_cifar_xnoise_sp_a}
    \end{subfigure}
    \hfill
    \begin{subfigure}[t]{0.49\linewidth}
        \centering
        \includegraphics[width=\linewidth]{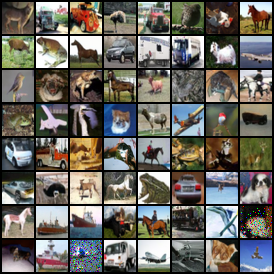}
        \caption{FM-TRS}
        \label{fig:app_cifar_xnoise_sp_b}
    \end{subfigure}
    \caption{\textbf{Additional CIFAR-10 target corruption samples under SP corruption.} Standard OT-FM reproduces visible corruption artifacts, while FM-TRS suppresses them and generates cleaner samples.}
    \label{fig:app_cifar_xnoise_sp}
\end{figure}

\begin{figure}[t!]
    \centering
    \begin{subfigure}[t]{0.49\linewidth}
        \centering
        \includegraphics[width=\linewidth]{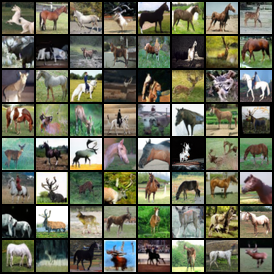}
        \caption{OT-FM}
        \label{fig:app_cifar_horse_a}
    \end{subfigure}
    \hfill
    \begin{subfigure}[t]{0.49\linewidth}
        \centering
        \includegraphics[width=\linewidth]{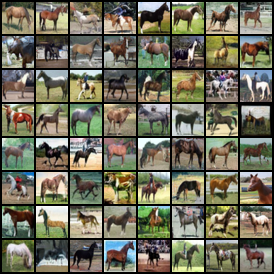}
        \caption{FM-CCS}
        \label{fig:app_cifar_horse_b}
    \end{subfigure}
    \caption{\textbf{Additional CIFAR-10 label-noise samples for the horse class under asymmetric $40\%$ noise.} Standard OT-FM often exhibits class confusion, whereas FM-CCS produces samples that are more consistently aligned with the requested class.}
    \label{fig:app_cifar_horse_asym04}
\end{figure}

Figures~\ref{fig:app_cifar_xnoise_sp} and~\ref{fig:app_cifar_horse_asym04} show representative CIFAR-10 generations under both noise corruptions.

Figure~\ref{fig:app_cifar_xnoise_sp} presents SP target corruption. Standard OT-FM memorizes visible corruption artifacts and produces samples with noticeable impulse-noise patterns. FM-TRS suppresses these artifacts and generates substantially cleaner images while preserving semantic diversity.

Figure~\ref{fig:app_cifar_horse_asym04} presents the horse class under asymmetric $40\%$ label noise. In this setting, the baseline often drifts toward visually plausible but incorrect categories, reflecting the conditional-misalignment failure mode discussed in the main text. FM-CCS improves class consistency and produces samples that are more reliably aligned with the requested horse condition.

\subsection{WebVision qualitative results}
\label{app:qual_webvision}

\begin{figure}[t]
    \centering
    \begin{subfigure}[t]{0.49\linewidth}
        \centering
        \includegraphics[width=\linewidth]{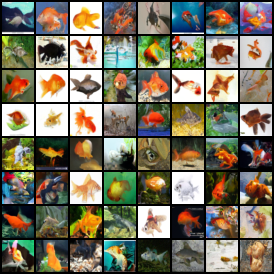}
        \caption{OT-FM}
        \label{fig:app_webvision_goldfish_a}
    \end{subfigure}
    \hfill
    \begin{subfigure}[t]{0.49\linewidth}
        \centering
        \includegraphics[width=\linewidth]{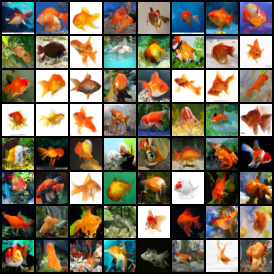}
        \caption{FM-CCS}
        \label{fig:app_webvision_goldfish_b}
    \end{subfigure}
    \caption{\textbf{Additional WebVision qualitative results for the goldfish class.} Under real-world noisy labels, standard OT-FM can produce visually plausible but weakly class-aligned samples. FM-CCS produces samples that are more consistently aligned with the requested class, indicating improved conditional robustness under naturally noisy supervision.}
    \label{fig:app_webvision_goldfish}
\end{figure}

Figure~\ref{fig:app_webvision_goldfish} provides a representative qualitative example on WebVision for the \emph{goldfish} class. In this example, standard OT-FM shows weaker class consistency and more visible drift toward ambiguous or weakly aligned visual content. In contrast, FM-CCS produces samples that are more consistently aligned with the requested class, which qualitatively matches the improvements in CAS, CW-FID, and CW-Coverage reported in Table~\ref{tab:webvision_results}.

\section{Scope and Limitations}
\label{app:limitations}

\paragraph{Residual reliability under target corruption.}
TRS is intended for settings in which corrupted targets form a non-dominating component and produce intermediate-to-late-time residuals distinguishable from those of the clean majority. Rare modes or legitimate high-frequency structure may receive large residuals and be downweighted, while corrupted targets close to the source distribution or otherwise easy to transport may retain high weights. Residual information alone cannot reliably distinguish these cases when their score distributions overlap. The diagnostics in Appendix~\ref{app:diagnostics} show score separation in the evaluated settings. Addressing these boundary cases may require additional information about the corruption process, assumptions about clean-data structure, or auxiliary reliability signals.

\paragraph{Temporal weighting in TRS.}
For computational efficiency, FM-TRS reuses the residual evaluated at the same late-biased time for score accumulation and optimization. It therefore optimizes a temporally reweighted rather than the standard uniform-time FM objective. The clean-data ablations in Appendix~\ref{app:clean_time_sampling} show no degradation in the reported mean metrics, but the effect may depend on the task, model, probability path, and optimization settings.

\paragraph{Comparative reliability and design choices.}
CCS requires distinguishable comparative losses on the evaluated trajectories. This signal may be less reliable early in training or when labels remain ambiguous. Warmup, probe times, and refresh schedules are practical choices supported by the ablations rather than universal prescriptions. TRS also exhibits sensitivity to EMA decay, and different training setups may benefit from tuning.

\paragraph{Condition spaces and computational cost.}
CCS is developed for finite discrete labels. Exact refreshes scale linearly with the number of conditions, while practical cost also increases with dataset size and model complexity. Larger-scale applications may require sampled or top-$k$ alternatives, hierarchical screening, or partial refreshes. Extension to continuous or structured conditions remains future work.

\paragraph{Mixed corruption and broader evaluation.}
The joint-corruption experiment in Appendix~\ref{app:joint_corruption} provides initial evidence on a controlled 2D benchmark. Broader settings motivate further study of score interactions, relative calibration, and alternative combination rules. Blur, compression, and severe class imbalance remain outside the current evaluation. Controlled comparisons with diffusion-based methods also remain future work, requiring matched corruption settings, architectures, training budgets, and available information.

\clearpage
\section*{NeurIPS Paper Checklist}

\begin{enumerate}

\item {\bf Claims}
    \item[] Question: Do the main claims made in the abstract and introduction accurately reflect the paper's contributions and scope?
    \item[] Answer: \answerYes{} 
    \item[] Justification: We summarize the contribution of our work in the introduction and in the abstract.
    \item[] Guidelines:
    \begin{itemize}
        \item The answer \answerNA{} means that the abstract and introduction do not include the claims made in the paper.
        \item The abstract and/or introduction should clearly state the claims made, including the contributions made in the paper and important assumptions and limitations. A \answerNo{} or \answerNA{} answer to this question will not be perceived well by the reviewers. 
        \item The claims made should match theoretical and experimental results, and reflect how much the results can be expected to generalize to other settings. 
        \item It is fine to include aspirational goals as motivation as long as it is clear that these goals are not attained by the paper. 
    \end{itemize}

\item {\bf Limitations}
    \item[] Question: Does the paper discuss the limitations of the work performed by the authors?
    \item[] Answer: \answerYes{} 
    \item[] Justification: TThe conclusion and Appendix~\ref{app:limitations} discuss residual reliability, difficult valid samples and easily fitted corruptions, temporal weighting in TRS, conditional-loss separation and computational scaling in CCS, and the scope of the joint-corruption evaluation.
    \item[] Guidelines:
    \begin{itemize}
        \item The answer \answerNA{} means that the paper has no limitation while the answer \answerNo{} means that the paper has limitations, but those are not discussed in the paper. 
        \item The authors are encouraged to create a separate ``Limitations'' section in their paper.
        \item The paper should point out any strong assumptions and how robust the results are to violations of these assumptions (e.g., independence assumptions, noiseless settings, model well-specification, asymptotic approximations only holding locally). The authors should reflect on how these assumptions might be violated in practice and what the implications would be.
        \item The authors should reflect on the scope of the claims made, e.g., if the approach was only tested on a few datasets or with a few runs. In general, empirical results often depend on implicit assumptions, which should be articulated.
        \item The authors should reflect on the factors that influence the performance of the approach. For example, a facial recognition algorithm may perform poorly when image resolution is low or images are taken in low lighting. Or a speech-to-text system might not be used reliably to provide closed captions for online lectures because it fails to handle technical jargon.
        \item The authors should discuss the computational efficiency of the proposed algorithms and how they scale with dataset size.
        \item If applicable, the authors should discuss possible limitations of their approach to address problems of privacy and fairness.
        \item While the authors might fear that complete honesty about limitations might be used by reviewers as grounds for rejection, a worse outcome might be that reviewers discover limitations that aren't acknowledged in the paper. The authors should use their best judgment and recognize that individual actions in favor of transparency play an important role in developing norms that preserve the integrity of the community. Reviewers will be specifically instructed to not penalize honesty concerning limitations.
    \end{itemize}

\item {\bf Theory assumptions and proofs}
    \item[] Question: For each theoretical result, does the paper provide the full set of assumptions and a complete (and correct) proof?
    \item[] Answer: \answerYes{} 
    \item[] Justification: The assumptions for the FM setup and corruption models are stated in Section~\ref{sec:problem}. The derivations for Propositions~\ref{prop:x_noise_decomposition} and~\ref{prop:y_noise_decomposition}, and the proof of the CCS sign-separation result, are provided in Appendix~\ref{app:proofs}.
    \item[] Guidelines:
    \begin{itemize}
        \item The answer \answerNA{} means that the paper does not include theoretical results. 
        \item All the theorems, formulas, and proofs in the paper should be numbered and cross-referenced.
        \item All assumptions should be clearly stated or referenced in the statement of any theorems.
        \item The proofs can either appear in the main paper or the supplemental material, but if they appear in the supplemental material, the authors are encouraged to provide a short proof sketch to provide intuition. 
        \item Inversely, any informal proof provided in the core of the paper should be complemented by formal proofs provided in appendix or supplemental material.
        \item Theorems and Lemmas that the proof relies upon should be properly referenced. 
    \end{itemize}

    \item {\bf Experimental result reproducibility}
    \item[] Question: Does the paper fully disclose all the information needed to reproduce the main experimental results of the paper to the extent that it affects the main claims and/or conclusions of the paper (regardless of whether the code and data are provided or not)?
    \item[] Answer: \answerYes{} 
    \item[] Justification: The paper specifies the datasets, corruption protocols, architectures, optimizers, training budgets, evaluation metrics, and baseline adaptations in Section~\ref{sec:experiments} and Appendix~\ref{app:implementation}. Pseudocode for both proposed methods is provided in Appendix~\ref{app:pseudocode}.
    \item[] Guidelines:
    \begin{itemize}
        \item The answer \answerNA{} means that the paper does not include experiments.
        \item If the paper includes experiments, a \answerNo{} answer to this question will not be perceived well by the reviewers: Making the paper reproducible is important, regardless of whether the code and data are provided or not.
        \item If the contribution is a dataset and\slash or model, the authors should describe the steps taken to make their results reproducible or verifiable. 
        \item Depending on the contribution, reproducibility can be accomplished in various ways. For example, if the contribution is a novel architecture, describing the architecture fully might suffice, or if the contribution is a specific model and empirical evaluation, it may be necessary to either make it possible for others to replicate the model with the same dataset, or provide access to the model. In general. releasing code and data is often one good way to accomplish this, but reproducibility can also be provided via detailed instructions for how to replicate the results, access to a hosted model (e.g., in the case of a large language model), releasing of a model checkpoint, or other means that are appropriate to the research performed.
        \item While NeurIPS does not require releasing code, the conference does require all submissions to provide some reasonable avenue for reproducibility, which may depend on the nature of the contribution. For example
        \begin{enumerate}
            \item If the contribution is primarily a new algorithm, the paper should make it clear how to reproduce that algorithm.
            \item If the contribution is primarily a new model architecture, the paper should describe the architecture clearly and fully.
            \item If the contribution is a new model (e.g., a large language model), then there should either be a way to access this model for reproducing the results or a way to reproduce the model (e.g., with an open-source dataset or instructions for how to construct the dataset).
            \item We recognize that reproducibility may be tricky in some cases, in which case authors are welcome to describe the particular way they provide for reproducibility. In the case of closed-source models, it may be that access to the model is limited in some way (e.g., to registered users), but it should be possible for other researchers to have some path to reproducing or verifying the results.
        \end{enumerate}
    \end{itemize}

\item {\bf Open access to data and code}
    \item[] Question: Does the paper provide open access to the data and code, with sufficient instructions to faithfully reproduce the main experimental results, as described in supplemental material?
    \item[] Answer: \answerYes{}
    \textbf{Justification:} The code for the proposed methods and the main experiments is publicly available at the repository linked in the paper. The repository provides setup instructions and scripts for reproducing the core 2D and CIFAR-10 target-corruption and label-noise experiments. The experiments use public benchmark datasets, and additional implementation details are provided in the appendix.
    \item[] Guidelines:
    \begin{itemize}
        \item The answer \answerNA{} means that paper does not include experiments requiring code.
        \item Please see the NeurIPS code and data submission guidelines (\url{https://neurips.cc/public/guides/CodeSubmissionPolicy}) for more details.
        \item While we encourage the release of code and data, we understand that this might not be possible, so \answerNo{} is an acceptable answer. Papers cannot be rejected simply for not including code, unless this is central to the contribution (e.g., for a new open-source benchmark).
        \item The instructions should contain the exact command and environment needed to run to reproduce the results. See the NeurIPS code and data submission guidelines (\url{https://neurips.cc/public/guides/CodeSubmissionPolicy}) for more details.
        \item The authors should provide instructions on data access and preparation, including how to access the raw data, preprocessed data, intermediate data, and generated data, etc.
        \item The authors should provide scripts to reproduce all experimental results for the new proposed method and baselines. If only a subset of experiments are reproducible, they should state which ones are omitted from the script and why.
        \item At submission time, to preserve anonymity, the authors should release anonymized versions (if applicable).
        \item Providing as much information as possible in supplemental material (appended to the paper) is recommended, but including URLs to data and code is permitted.
    \end{itemize}

\item {\bf Experimental setting/details}
    \item[] Question: Does the paper specify all the training and test details (e.g., data splits, hyperparameters, how they were chosen, type of optimizer) necessary to understand the results?
    \item[] Answer: \answerYes{} 
    \item[] Justification: Training details, architectures, optimizers, corruption models, hyperparameters, sampling settings, and evaluation protocols are described in Appendix~\ref{app:implementation}. The main experimental setup is summarized in Section~\ref{sec:experiments}.
    \item[] Guidelines:
    \begin{itemize}
        \item The answer \answerNA{} means that the paper does not include experiments.
        \item The experimental setting should be presented in the core of the paper to a level of detail that is necessary to appreciate the results and make sense of them.
        \item The full details can be provided either with the code, in appendix, or as supplemental material.
    \end{itemize}

\item {\bf Experiment statistical significance}
    \item[] Question: Does the paper report error bars suitably and correctly defined or other appropriate information about the statistical significance of the experiments?
    \item[] Answer: \answerYes{} 
    \item[] Justification: For every method, paired seed-wise significance tests are reported in Appendix~\ref{app:statistical_significance}.
    \item[] Guidelines:
    \begin{itemize}
        \item The answer \answerNA{} means that the paper does not include experiments.
        \item The authors should answer \answerYes{} if the results are accompanied by error bars, confidence intervals, or statistical significance tests, at least for the experiments that support the main claims of the paper.
        \item The factors of variability that the error bars are capturing should be clearly stated (for example, train/test split, initialization, random drawing of some parameter, or overall run with given experimental conditions).
        \item The method for calculating the error bars should be explained (closed form formula, call to a library function, bootstrap, etc.)
        \item The assumptions made should be given (e.g., Normally distributed errors).
        \item It should be clear whether the error bar is the standard deviation or the standard error of the mean.
        \item It is OK to report 1-sigma error bars, but one should state it. The authors should preferably report a 2-sigma error bar than state that they have a 96\% CI, if the hypothesis of Normality of errors is not verified.
        \item For asymmetric distributions, the authors should be careful not to show in tables or figures symmetric error bars that would yield results that are out of range (e.g., negative error rates).
        \item If error bars are reported in tables or plots, the authors should explain in the text how they were calculated and reference the corresponding figures or tables in the text.
    \end{itemize}

\item {\bf Experiments compute resources}
    \item[] Question: For each experiment, does the paper provide sufficient information on the computer resources (type of compute workers, memory, time of execution) needed to reproduce the experiments?
    \item[] Answer: \answerYes{} 
    \item[] Justification: Appendix~\ref{app:computational_cost} reports measured wall-clock training times and relative overheads for the main image-scale experiments. The appendix also specifies the compute workers used for the experiments and the training budgets needed to reproduce the reported runs.
    \item[] Guidelines:
    \begin{itemize}
        \item The answer \answerNA{} means that the paper does not include experiments.
        \item The paper should indicate the type of compute workers CPU or GPU, internal cluster, or cloud provider, including relevant memory and storage.
        \item The paper should provide the amount of compute required for each of the individual experimental runs as well as estimate the total compute. 
        \item The paper should disclose whether the full research project required more compute than the experiments reported in the paper (e.g., preliminary or failed experiments that didn't make it into the paper). 
    \end{itemize}
    
\item {\bf Code of ethics}
    \item[] Question: Does the research conducted in the paper conform, in every respect, with the NeurIPS Code of Ethics \url{https://neurips.cc/public/EthicsGuidelines}?
    \item[] Answer: \answerYes{} 
    \item[] Justification: The work uses standard public datasets and proposes training methods for robust generative modeling. We are not aware of any aspect of the work that violates the NeurIPS Code of Ethics.
    \item[] Guidelines:
    \begin{itemize}
        \item The answer \answerNA{} means that the authors have not reviewed the NeurIPS Code of Ethics.
        \item If the authors answer \answerNo, they should explain the special circumstances that require a deviation from the Code of Ethics.
        \item The authors should make sure to preserve anonymity (e.g., if there is a special consideration due to laws or regulations in their jurisdiction).
    \end{itemize}

\item {\bf Broader impacts}
    \item[] Question: Does the paper discuss both potential positive societal impacts and negative societal impacts of the work performed?
    \item[] Answer: \answerNA{}.
    \item[] Justification: This work is primarily methodological and is not tied to a specific deployed application. Its potential positive impact is to improve the robustness of flow matching models trained with corrupted targets or noisy labels, which may support more reliable generative modeling in research settings. At the same time, the methods inherit general dual-use concerns associated with improved generative models, including possible misuse in image generation.
    \item[] Guidelines:
    \begin{itemize}
        \item The answer \answerNA{} means that there is no societal impact of the work performed.
        \item If the authors answer \answerNA{} or \answerNo, they should explain why their work has no societal impact or why the paper does not address societal impact.
        \item Examples of negative societal impacts include potential malicious or unintended uses (e.g., disinformation, generating fake profiles, surveillance), fairness considerations (e.g., deployment of technologies that could make decisions that unfairly impact specific groups), privacy considerations, and security considerations.
        \item The conference expects that many papers will be foundational research and not tied to particular applications, let alone deployments. However, if there is a direct path to any negative applications, the authors should point it out. For example, it is legitimate to point out that an improvement in the quality of generative models could be used to generate Deepfakes for disinformation. On the other hand, it is not needed to point out that a generic algorithm for optimizing neural networks could enable people to train models that generate Deepfakes faster.
        \item The authors should consider possible harms that could arise when the technology is being used as intended and functioning correctly, harms that could arise when the technology is being used as intended but gives incorrect results, and harms following from (intentional or unintentional) misuse of the technology.
        \item If there are negative societal impacts, the authors could also discuss possible mitigation strategies (e.g., gated release of models, providing defenses in addition to attacks, mechanisms for monitoring misuse, mechanisms to monitor how a system learns from feedback over time, improving the efficiency and accessibility of ML).
    \end{itemize}
    
\item {\bf Safeguards}
    \item[] Question: Does the paper describe safeguards that have been put in place for responsible release of data or models that have a high risk for misuse (e.g., pre-trained language models, image generators, or scraped datasets)?
    \item[] Answer: \answerNA{} 
    \item[] Justification: The paper does not release a high-risk pretrained generative model or a new scraped dataset. Experiments are performed on existing benchmark datasets.
    \item[] Guidelines:
    \begin{itemize}
        \item The answer \answerNA{} means that the paper poses no such risks.
        \item Released models that have a high risk for misuse or dual-use should be released with necessary safeguards to allow for controlled use of the model, for example by requiring that users adhere to usage guidelines or restrictions to access the model or implementing safety filters. 
        \item Datasets that have been scraped from the Internet could pose safety risks. The authors should describe how they avoided releasing unsafe images.
        \item We recognize that providing effective safeguards is challenging, and many papers do not require this, but we encourage authors to take this into account and make a best faith effort.
    \end{itemize}

\item {\bf Licenses for existing assets}
    \item[] Question: Are the creators or original owners of assets (e.g., code, data, models), used in the paper, properly credited and are the license and terms of use explicitly mentioned and properly respected?
    \item[] Answer: \answerYes{} 
    \item[] Justification: We properly cite all research works we build on and use the code of others only according to its license.
    \item[] Guidelines:
    \begin{itemize}
        \item The answer \answerNA{} means that the paper does not use existing assets.
        \item The authors should cite the original paper that produced the code package or dataset.
        \item The authors should state which version of the asset is used and, if possible, include a URL.
        \item The name of the license (e.g., CC-BY 4.0) should be included for each asset.
        \item For scraped data from a particular source (e.g., website), the copyright and terms of service of that source should be provided.
        \item If assets are released, the license, copyright information, and terms of use in the package should be provided. For popular datasets, \url{paperswithcode.com/datasets} has curated licenses for some datasets. Their licensing guide can help determine the license of a dataset.
        \item For existing datasets that are re-packaged, both the original license and the license of the derived asset (if it has changed) should be provided.
        \item If this information is not available online, the authors are encouraged to reach out to the asset's creators.
    \end{itemize}

\item {\bf New assets}
    \item[] Question: Are new assets introduced in the paper well documented and is the documentation provided alongside the assets?
    \item[] Answer: \answerNA{} 
    \item[] Justification: The paper does not introduce a new dataset, benchmark, pretrained model release, or other standalone asset.
    \item[] Guidelines:
    \begin{itemize}
        \item The answer \answerNA{} means that the paper does not release new assets.
        \item Researchers should communicate the details of the dataset\slash code\slash model as part of their submissions via structured templates. This includes details about training, license, limitations, etc. 
        \item The paper should discuss whether and how consent was obtained from people whose asset is used.
        \item At submission time, remember to anonymize your assets (if applicable). You can either create an anonymized URL or include an anonymized zip file.
    \end{itemize}

\item {\bf Crowdsourcing and research with human subjects}
    \item[] Question: For crowdsourcing experiments and research with human subjects, does the paper include the full text of instructions given to participants and screenshots, if applicable, as well as details about compensation (if any)? 
    \item[] Answer: \answerNA{} 
    \item[] Justification: The paper does not involve crowdsourcing experiments or research with human subjects.
    \item[] Guidelines:
    \begin{itemize}
        \item The answer \answerNA{} means that the paper does not involve crowdsourcing nor research with human subjects.
        \item Including this information in the supplemental material is fine, but if the main contribution of the paper involves human subjects, then as much detail as possible should be included in the main paper. 
        \item According to the NeurIPS Code of Ethics, workers involved in data collection, curation, or other labor should be paid at least the minimum wage in the country of the data collector. 
    \end{itemize}

\item {\bf Institutional review board (IRB) approvals or equivalent for research with human subjects}
    \item[] Question: Does the paper describe potential risks incurred by study participants, whether such risks were disclosed to the subjects, and whether Institutional Review Board (IRB) approvals (or an equivalent approval/review based on the requirements of your country or institution) were obtained?
    \item[] Answer: \answerNA{} 
    \item[] Justification: The paper does not involve crowdsourcing experiments or research with human subjects.
    \item[] Guidelines:
    \begin{itemize}
        \item The answer \answerNA{} means that the paper does not involve crowdsourcing nor research with human subjects.
        \item Depending on the country in which research is conducted, IRB approval (or equivalent) may be required for any human subjects research. If you obtained IRB approval, you should clearly state this in the paper. 
        \item We recognize that the procedures for this may vary significantly between institutions and locations, and we expect authors to adhere to the NeurIPS Code of Ethics and the guidelines for their institution. 
        \item For initial submissions, do not include any information that would break anonymity (if applicable), such as the institution conducting the review.
    \end{itemize}

\item {\bf Declaration of LLM usage}
    \item[] Question: Does the paper describe the usage of LLMs if it is an important, original, or non-standard component of the core methods in this research? Note that if the LLM is used only for writing, editing, or formatting purposes and does \emph{not} impact the core methodology, scientific rigor, or originality of the research, declaration is not required.
    \item[] Answer: \answerNA{} 
    \item[] Justification: LLMs are not used as an important, original, or non-standard component of the core methodology.
    \item[] Guidelines:
    \begin{itemize}
        \item The answer \answerNA{} means that the core method development in this research does not involve LLMs as any important, original, or non-standard components.
        \item Please refer to our LLM policy in the NeurIPS handbook for what should or should not be described.
    \end{itemize}

\end{enumerate}

\end{document}